\documentclass[a4paper,12pt]{article}

\usepackage[T1]{fontenc}
\usepackage{tabularray}
\usepackage{times}
\usepackage{amssymb}
\usepackage[english]{babel}
\usepackage[utf8]{inputenc}
\usepackage{dtklogos}
\usepackage{wallpaper}
\usepackage[absolute]{textpos}
\usepackage[top=2cm, bottom=2.5cm, left=3cm, right=3cm]{geometry}
\usepackage{appendix}
\usepackage[nottoc]{tocbibind}
\usepackage[colorlinks=true,
            linkcolor=black,
            urlcolor=blue,
            citecolor=black]{hyperref}
\usepackage{csvsimple}
\usepackage{graphicx}
\usepackage{tabularx}
\usepackage{multirow}
\usepackage{makecell}
\usepackage{array}
\usepackage{booktabs} % For better looking tables
\usepackage{arydshln} % For dashed lines in tables
\usepackage{tikz}
\usepackage{pgfplots}
\usepackage{rotating}
\usepackage{caption}
\usepackage{booktabs}
\usepackage{xurl} % Allows line breaks in URLs/DOIs if they are long
\usepackage{doi}  % Provides a \doi{} command that hyperref can often link
\usepackage[normalem]{ulem} % allows underlining and overstriking
\usepackage{siunitx}
\usepackage{amsmath}
\pgfplotsset{compat=1.17}

\usepackage{sectsty}
\sectionfont{\fontsize{14}{15}\selectfont}
\subsectionfont{\fontsize{12}{15}\selectfont}
\subsubsectionfont{\fontsize{12}{15}\selectfont}

\usepackage{csquotes} % Used to handle citations

\newsavebox{\mybox}
\newlength{\mydepth}
\newlength{\myheight}

\newenvironment{sidebar}%
{\begin{lrbox}{\mybox}\begin{minipage}{\textwidth}}%
{\end{minipage}\end{lrbox}%
 \settodepth{\mydepth}{\usebox{\mybox}}%
 \settoheight{\myheight}{\usebox{\mybox}}%
 \addtolength{\myheight}{\mydepth}%
 \noindent\makebox[0pt]{\hspace{-20pt}\rule[-\mydepth]{1pt}{\myheight}}%
 \usebox{\mybox}}

\newcommand\BackgroundPic{
    \put(-2,-3){
    \includegraphics[keepaspectratio,scale=0.3]{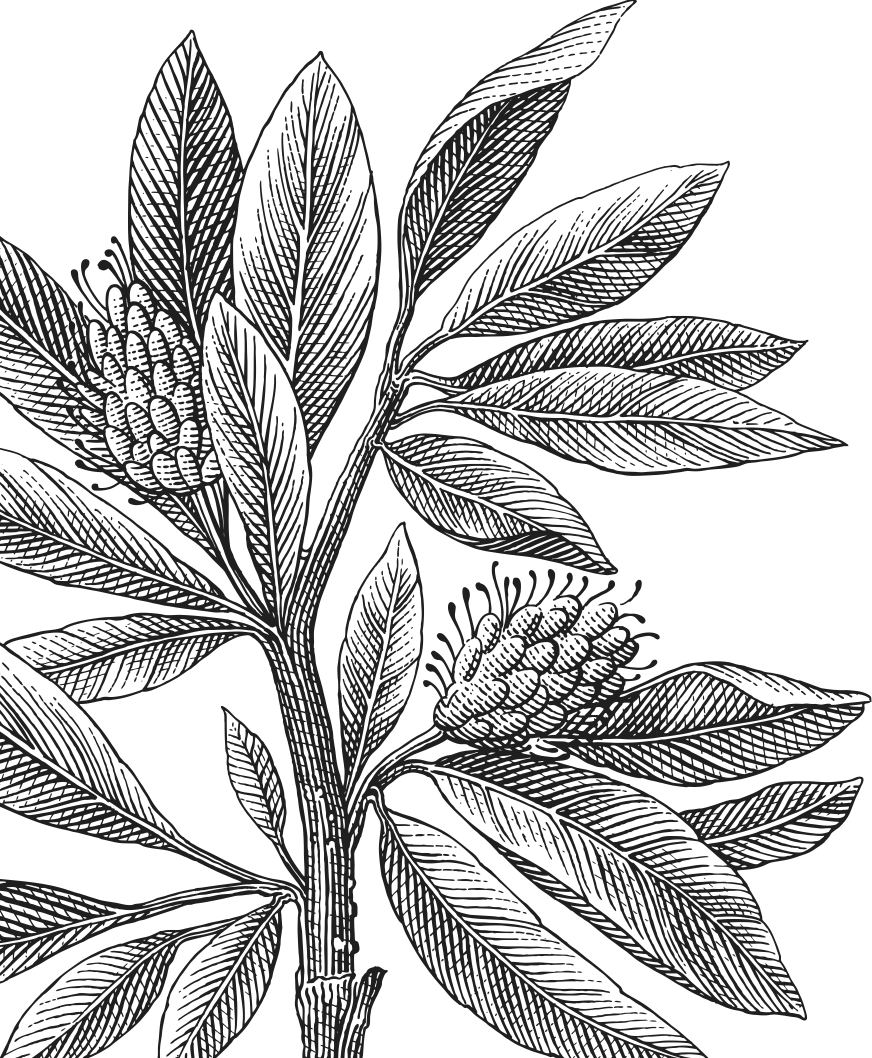} % Background picture
    }
}
\newcommand\BackgroundPicLogo{
    \put(30,740){
    \includegraphics[keepaspectratio,scale=0.10]{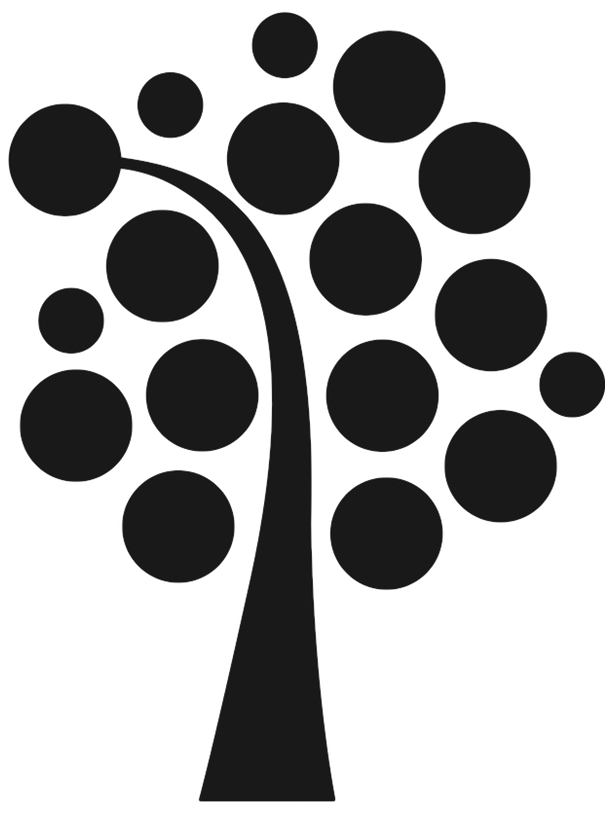} % Logo in upper left corner
    }
}

\title{	
\vspace{-8cm}
\begin{sidebar}
    \vspace{10cm}
    \normalfont \normalsize
    \Huge Masters Degree Project \\
    \vspace{-1.3cm}
\end{sidebar}
\vspace{3cm}
\begin{flushleft}
    A Systematic Approach for Selecting Trajectories for Data Augmentation
\end{flushleft}
\null
\vfill
\begin{textblock}{6}(10,13)
\begin{flushright}
\begin{minipage}{\textwidth}
\begin{flushleft} \normalsize
\begin{tabular}{p{0.4pt}|p{6cm}}
  & \emph{Author:}  Adam Nordling\\ % Author
  & \emph{Supervisor:} Amilcar Soares \\ % Supervisor
  & \emph{Examiner:} Hemant Ghayvat  \\ %Dr.~Daniel \textsc{Toll}\\ % Examiner (course manager)
  & \emph{Semester:} VT 2026\\ % 
  & \emph{Course:} 5DV50E \\ % Course
  & \emph{Subject:} Computer Science \\ % Subject area
\end{tabular}
\end{flushleft}
\end{minipage}
\end{flushright}
\end{textblock}
}

\date{}

\begin{document}
\pagenumbering{gobble}
\newgeometry{left=5cm}
\AddToShipoutPicture*{\BackgroundPic}
\AddToShipoutPicture*{\BackgroundPicLogo}
\maketitle
\restoregeometry
\clearpage
%----------------------------------------------------------------------------------------
%	Abstract
%----------------------------------------------------------------------------------------
\selectlanguage{english}

\begin{abstract}
\noindent Trajectory data augmentation is a promising approach to mitigate data scarcity in machine learning applications, but its utility has been limited by the complexity of preserving spatio-temporal coherence. Although prior work demonstrated the viability of geometric perturbation, it relied on naive random selection, leaving a critical gap in understanding \textit{which} trajectories should be augmented for maximal benefit. This thesis addresses this gap by developing a systematic and scalable framework to evaluate five systematic selection strategies: Outlierness, Diversity, Representativeness, Uncertainty, and Random selection. 
These strategies were rigorously tested across four datasets covering animal behavior (Foxes and Starkey), maritime traffic (AIS), and urban traffic (Car) using a suite of linear and non-linear machine learning models. As part of this evaluation, an Optuna-based hyperparameter optimization loop was integrated to empirically identify the best-performing augmentation parameters for each dataset within the explored search space.
The results indicate that, while systematic selection is not a universal solution, it offers distinct advantages over the random baseline. 
Systematic strategies, particularly Outlierness and Uncertainty, demonstrated higher stability and were less prone to performance degradation observed with random sampling in dense datasets. However, the findings also reveal that the value of augmentation is strictly conditional. 
Visual analysis via UMAP demonstrates that while systematic augmentation successfully repairs topological fragmentation in sparse datasets, it can act as a corrupting noise signal in high-quality, dense datasets. Furthermore, the study identified physical limitations in high-velocity domains, where standard perturbation techniques lead to divergence in feature space. 
This research provides a practical evidence-based framework for practitioners, establishing that strategic data selection is not only an optimization detail, but can be the primary determinant of augmentation success.
\end{abstract}

\newpage
%----------------------------------------------------------------------------------------
%	Preface
%----------------------------------------------------------------------------------------

\textbf {\large{Preface}}\\

\noindent This thesis investigates systematic selection strategies for augmentation of trajectory data to improve machine learning classification. The research topic was initiated by my supervisor, Amilcar Soares, as a refinement of future work suggested in previous studies within this field. \newline

\noindent I would like to thank Amilcar Soares for his guidance throughout the entire project, from the initial formulation of the research question to the final results analysis. His expertise was essential in navigating the complexities of trajectory data and ensured that the findings were interpreted with the necessary scientific nuance. \newline

\noindent In the production of this thesis, some artificial intelligence tools were used to help solve specific programming challenges and to improve the clarity of written explanations. Although these tools provided technical and linguistic support, all conceptual decisions and final conclusions remain my own independent work.

%----------------------------------------------------------------------------------------
\newpage
\pagenumbering{gobble}
\tableofcontents % Table of contents
\newpage
\pagenumbering{arabic}

%----------------------------------------------------------------------------------------
%
%	Here follows the actual text contents of the report.
%
%----------------------------------------------------------------------------------------
\section{Introduction}

\subsection{Background}

Trajectories represent the movement of entities, such as people, vehicles, or animals, through space and over time. 
Typically modeled as a sequence of timestamped geographic coordinates, trajectories have become central to a wide range of applications, including urban planning \cite{zhao2023survey}, transportation optimization \cite{liu2024transport}, wildlife monitoring \cite{garrido2023gnss}, and maritime traffic analysis \cite{ribeiro2020survey, abreu2021trajectory}. 
For example, the data collected from the Automatic Identification System (AIS) supports the monitoring of maritime traffic networks \cite{song2024enhancing, carlini2022understanding,soares2019crisis} and enables the forecast of vessel routes \cite{spadon2024multi}. 
In fisheries management, vessel trajectories help detect fishing activity \cite{ferreira2022semi,adibi2019predicting}, while in smart mobility and urban analytics, trajectory data plays a key role in understanding movement patterns in cities \cite{haranwala2022dashboard}.

Despite their utility, trajectory datasets pose substantial analytical challenges. 
Trajectories are inherently spatiotemporal and often sampled irregularly, making them difficult to preprocess and model \cite{haidri2022ptrail}. 
Additional complications arise from GPS measurement noise, semantic ambiguities in movement patterns, and the high dimensionality of derived features. 
Moreover, privacy concerns and proprietary data ownership frequently prevent the public release of high-quality trajectory datasets, limiting reproducibility and benchmarking in research \cite{haranwala2023data}. 
These difficulties are compounded in large-scale settings where big data techniques are necessary to handle large volumes of movement records \cite{ribeiro2020survey, ferreira2023assessing}.

Trajectory data supports a wide range of tasks, such as trajectory classification \cite{lee2023trajectory,junior2017analytic, tavakoli2022study}, clustering \cite{wang2023trajectory,johnson2023hail}, anomaly detection \cite{lan2023anomaly,abreu2021trajectory}, segmentation \cite{tan2023motion, etemad2021sws, junior2018semi, soares2015grasp}, and forecasting \cite{spadon2022unfolding,may2020challenges}. 
For example, detecting unusual vessel behavior requires analyzing deviations from normal movement patterns using spatial, temporal, and contextual cues \cite{carlini2020uncovering}. 
Researchers are increasingly applying machine learning and deep learning methods to these tasks, but these models often rely on large labeled datasets for training and evaluation. These datasets are typically scarce or incomplete in this domain.

One promising approach to mitigate labeled data scarcity is trajectory data augmentation, which artificially expands the training dataset by generating new realistic examples from existing data. 
Augmentation techniques are widely adopted in computer vision and natural language processing, where simple transformations (e.g., rotations or word substitutions) can improve model generalization and reduce overfitting. 
However, in the trajectory domain, augmentation remains underexplored \cite{haranwala2023data}. 
This is in part due to the complexity of maintaining the spatiotemporal coherence and semantic plausibility of generated trajectories. 
Designing transformations that preserve key characteristics (such as speed, direction, and behavioral patterns) is a non-trivial task.

\subsection{Motivation}

Trajectory data augmentation presents several compelling opportunities in the context of movement analysis. 
First, it helps mitigate the common challenge of data scarcity in trajectory-based applications, such as urban planning, transportation optimization, wildlife monitoring, and maritime traffic analysis, by synthetically increasing the availability of labeled data. 
This enables the development of more robust and generalizable machine learning models. Second, augmentation strategies can help address class imbalance problems frequently observed in trajectory datasets, where certain types of movement, behaviors, or transportation modes are underrepresented. 
By enriching these underrepresented categories, augmentation supports more equitable and effective model training. 
Finally, systematic trajectory augmentation can reveal which geometric or temporal features are most discriminative for downstream tasks such as trajectory classification, clustering, or anomaly detection. 
This contributes not only to improved model performance but also to a deeper understanding of the structure and variability of real-world mobility patterns.

\subsection{Problem Formulation}

Haranwala et al.~\cite{haranwala2023data} proposed a trajectory augmentation algorithm that introduces geographical perturbations to selected trajectory points using two strategies: in-circle and on-circle noise. 
The method randomly selects points along a trajectory and modifies their geographic coordinates within predefined circular regions, generating augmented trajectories that preserve key spatio-temporal properties while introducing controlled variability. %Their results showed notable improvements—approximately 20%—in trajectory classification accuracy across three benchmark datasets.

However, as the authors themselves acknowledged, their implementation relied entirely on the random selection of both trajectories and trajectory points for noise injection. 
Although this randomness effectively demonstrated the viability of their approach, it also limited its applicability and optimization potential. 
The absence of a principled selection mechanism raises an important open question in the field:

\begin{center}
\textit{How can trajectories be systematically selected for augmentation to improve classification performance?}
\end{center}

\noindent This research question focuses on a gap identified by Haranwala et al.~\cite{haranwala2023data}, who emphasized the need for “systematically controlling this noise to enhance its applicability in classification problems” and noted that “innovating new techniques for ranking trajectories presents an exciting possibility for augmenting trajectories.”

\subsection{Objectives}

This thesis seeks to advance trajectory data augmentation by moving beyond random sampling toward principled methods for selecting trajectories and trajectory points to augment. 
The goal is to make augmentation more effective, interpretable and tailored to downstream classification tasks. \\

To address the research question, the thesis pursues the following objectives: \\

\begin{tabular}{|>{\centering\arraybackslash}m{1.2cm}|m{11.6cm}|} \hline
\textbf{O1} & Design and implement systematic selection strategies for trajectory augmentation, leveraging properties such as outlierness, representativeness, diversity, and prediction uncertainty.  \\ \hline
\textbf{O2} & Quantitatively evaluate the impact of different selection strategies on classification performance across multiple datasets and model architectures. \\ \hline
\textbf{O3} & Develop a framework for benchmarking trajectory selection strategies, enabling comparative analysis and providing empirical evidence for their effectiveness in augmentation tasks. \\ \hline
\end{tabular}

\newpage

\subsection{Contributions of the Work}

This thesis makes the following contributions to the field of trajectory data analysis:

\begin{itemize}
    \item \textbf{Theoretical Contributions:} It establishes theoretical foundations for systematically selecting trajectories and trajectory points. By doing so, this work provides the principled, data-driven framework that was absent from the purely random methods of previous research.
    % This contribution addresses a specific research gap identified by Haranwala et al. \cite{haranwala2023data} in their future work discussion.
    
    \item \textbf{Empirical Contributions:} It provides rigorous empirical evidence on which selection strategies yield the greatest improvement in classification performance. 
    This offers practical evidence-based guidelines that help researchers and practitioners make informed decisions.
    
    \item \textbf{Technical Contributions:} It delivers a high-performance, modular trajectory augmentation framework optimized for scalability. By implementing vectorized operations and efficient binary serialization, the system goes beyond single-use scripts to a comprehensive solution that integrates systematic selection with proven noise techniques at scale. 
    The complete software framework has been archived and made publicly available to ensure reproducibility~\cite{nordling_2025_code}.
\end{itemize}

\subsection{Scope and Limitations}

This research focuses specifically on extending the trajectory augmentation algorithm proposed by Haranwala et al.~\cite{haranwala2023data} by developing systematic selection strategies.
Although the original geographical noise techniques (specifically on-circle) will be utilized, this work will not explore alternative noise application methods. 
The evaluation uses trajectory datasets with spatial representations and preprocessing assumptions comparable to those in the original study. 
In addition to one overlapping dataset, additional datasets obtained through the same sources are included to support a consistent evaluation.

The thesis will focus on trajectory classification rather than regression or clustering. Additionally, the work will employ a diverse suite of machine learning models, ranging from linear to complex non-linear architectures, to enable robust evaluation of how different selection strategies impact model performance.

\subsection{Target Audience}

This research aims at three primary audiences: (1) researchers in mobility data analysis looking to overcome data scarcity challenges, (2) data scientists and machine learning practitioners working with trajectory data across various domains, and (3) domain experts in transportation, urban planning and location-based services who can benefit from improved trajectory classification models.

\subsection{Thesis Structure}

The remainder of this thesis is organized as follows:

\begin{itemize}
\item \textbf{Chapter 2 - Related Work:} Reviews relevant literature on trajectory data analysis, augmentation techniques in spatiotemporal domains, and methods for identifying important elements in trajectory data.
\item \textbf{Chapter 3 - Methodology:} Details the proposed systematic selection strategies for trajectories and trajectory points, along with the experimental setup for evaluating their effectiveness.
\item \textbf{Chapter 4 - Results and Analysis:} Presents findings from experiments comparing systematic selection approaches against random selection across different datasets and models.
\item \textbf{Chapter 5 - Discussion:} Interprets results and discusses implications, limitations, and insights gained about trajectory characteristics and their importance for classification tasks.
\item \textbf{Chapter 6 - Conclusion and Future Work:} Summarizes key contributions and outlines promising directions for future research in trajectory data augmentation.
\end{itemize}

\newpage

\section{Related Work}
\label{chap:related_work}

This chapter provides an overview of existing research and previous work relevant to the challenges of data scarcity and the application of data augmentation techniques in trajectory analysis. 
The purpose of this section is to position the current study within the academic landscape, demonstrating how it is built upon or differs from previous research on trajectory data augmentation and selection strategies. It will summarize key contributions, compare different approaches, and identify the specific gaps that this thesis aims to address. 

\subsection{Review of Existing Research}
\label{sec:review_existing_research}

Data augmentation is a well-known set of techniques for artificially increasing the size and diversity of training datasets, thus improving the generalizability of machine learning models and mitigating issues such as overfitting \cite{chen2023empirical, wang2024mixsurvey}. 
Although widely used in fields such as computer vision \cite{xu2023imageaug} and natural language processing \cite{chen2023empirical}, its application to trajectory data has been relatively limited due to inherent spatiotemporal complexities and the need to preserve the semantic integrity of movement patterns.

The work of Haranwala et al.~\cite{haranwala2023data} introduced a promising approach for trajectory data augmentation. 
They proposed an algorithm that applies geographical perturbations to trajectory points using two main strategies: \enquote{in-circle} noise and \enquote{on-circle} noise. 
This method randomly selects trajectory points and modifies their geographic coordinates within predefined circular regions, achieving significant performance improvements in trajectory classification across multiple datasets. 
However, a key acknowledgment in their work was the reliance on completely random selection of both trajectories and trajectory points for noise injection, highlighting a need for more systematic control. 

Building upon this concept, Haranwala's subsequent work introduced AugmenTRAJ, an open-source Python framework specifically designed for point-based trajectory data augmentation~\cite{haranwala2023augmenttraj}. 
AugmenTRAJ offers a suite of techniques, including several \textit{candidate trajectory selection strategies} such as random selection, proportional selection (based on class representation), length-based selection (focusing on shorter trajectories), and representative trajectory selection (based on statistical similarity to the overall dataset). It also expanded on \textit{point modification techniques}, including in-circle and on-circle modifications, point stretching, and point dropping. 
This framework provides a more structured approach to trajectory augmentation than purely random methods.

Beyond direct trajectory data augmentation for classification, related concepts appear in other specialized domains. 
For example, Agrawal et al.~\cite{agrawal2023trajaug} developed TrajAug for robust neural locomotion controllers, in which synthetic motion data are generated by motion matching to follow random trajectories. 
This approach aims to re-balance datasets and introduce sharper, more game-relevant turns, which are often underrepresented in captured motion data. Although the application (locomotion) and the generation method (motion matching) differ, the underlying principle of creating diverse, balanced data through synthetic trajectories remains relevant. 

In the field of imitation learning, Antotsiou et al.~\cite{antotsiou2021adversarial} addressed the challenge of preserving the validity of augmented trajectories. 
They introduced a semi-supervised correction network to rectify distorted expert actions, ensuring that augmented trajectories remain successful in control tasks. This is a critical factor, since random distortions can easily invalidate sequential decision-making processes.

For safety-critical applications such as autonomous driving, Mirkhani et al.~\cite{mirkhani2024augmenting} proposed a method to augment driving scenarios while preserving their similarity to expert trajectories. Their approach involves clustering trajectories to identify minority, safety-critical groups, combining trajectories within the same cluster using geometric transformations, and then performing rigorous safety and quality checks. This highlights the importance of context-aware augmentation, especially when dealing with high-stakes and imbalanced data. 

The problem of \textit{what} data to augment, rather than just \textit{how} to augment it, has been more explicitly addressed in other domains, such as computer vision. 
Lin et al.~\cite{lin2023selectaugment} developed SelectAugment, an approach that deterministically selects samples for augmentation online using Hierarchical Reinforcement Learning (HRL). SelectAugment decides whether to augment an image based on its content and the network's current training status, aiming to avoid potential negative effects of random augmentation, such as visual ambiguities or induced training biases. 

Ultimately, the goal of these augmentation strategies in this thesis is to improve trajectory classification. 
Surveys like the one by da Silva et al.~\cite{dasilva2019survey} review various trajectory classification methods, noting that the core challenge often lies in extracting or identifying features and subtrajectories that best discriminate between classes. 
Effective data augmentation can directly support this by providing more and more diverse data from which classifiers can learn. 

\subsection{Comparison of Approaches}
\label{sec:comparison_approaches}

The reviewed literature shows a range of approaches to data augmentation for trajectory data. The work by Haranwala et al.~\cite{haranwala2023data} established the effectiveness of point perturbation, applied to randomly selected trajectories. 
The AugmenTRAJ framework~\cite{haranwala2023augmenttraj} formalized this by presenting random selection as one of several explicit strategies (alongside proportional, length-based, etc.), giving users more control over the augmentation process. 
This thesis adopts this view, treating random selection as a baseline augmentation strategy, against which more complex data-driven methods can be compared. 

In contrast to these perturbation-based methods, approaches such as TrajAug by Agrawal et al.~\cite{agrawal2023trajaug} create entirely new trajectories using generative techniques such as motion matching. 
This represents a different philosophy, focused on synthesizing new data rather than modifying existing points. 

A key difference between methods is how they choose what to generate or select. 
Random selection offers simplicity and serves as an important benchmark for other augmentation techniques. 
The rule-based strategies in AugmenTRAJ~\cite{haranwala2023augmenttraj} provide alternative heuristic improvements. 
More advanced methods use learning or quality checks. 
For example, Antotsiou et al.~\cite{antotsiou2021adversarial} use a correction network to ensure that the augmented trajectories are still valid, while Mirkhani et al.~\cite{mirkhani2024augmenting} use clustering to find and enhance safety-critical driving scenarios. These methods focus on ensuring that the augmented data makes sense in its specific context. 

The work of Lin et al.~in SelectAugment~\cite{lin2023selectaugment}, though in the image domain, introduces the powerful idea of learning \textit{which} samples to augment. This data-driven approach, which considers the data itself, contrasts with fixed rules or a purely random choice, and serves as an inspiration for the systematic strategies we explore in this thesis.

Finally, the primary goal of these methods varies. Some, like Haranwala et al.~\cite{haranwala2023data} and AugmenTRAJ~\cite{haranwala2023augmenttraj}, aim to improve general trajectory classification. 
Others are built for specific tasks such as locomotion control (Agrawal et al.~\cite{agrawal2023trajaug}) or autonomous driving safety (Mirkhani et al.~\cite{mirkhani2024augmenting}).

\subsection{Identified Gaps and Positioning of this Work}
\label{sec:identified_gaps}

The existing literature, particularly Haranwala et al.~\cite{haranwala2023data}, clearly identifies a key limitation: the use of random selection in trajectory data augmentation. 
As they noted, \enquote{innovating new techniques for ranking trajectories presents an exciting possibility.} The AugmenTRAJ framework~\cite{haranwala2023augmenttraj} began to address this by offering several heuristic-based strategies. 
However, these heuristics are predefined and cannot be adaptable. 
The development of data-driven selection methods that consider the statistical properties of trajectories within their class (such as their outlierness, representativeness, contribution to diversity, or prediction uncertainty) remains an open area for investigation. This is the primary gap this thesis aims to fill.

Although sophisticated selection mechanisms such as SelectAugment~\cite{lin2023selectaugment} exist for image data, their application to the unique spatio-temporal nature of trajectory data has not been thoroughly explored. Trajectory data presents different challenges and opportunities for defining \enquote{sample content} and \enquote{training status} relevant for selection. 

This thesis directly addresses these gaps by moving beyond the random selection approach of Haranwala et al.~\cite{haranwala2023data} and the static heuristics in AugmenTRAJ~\cite{haranwala2023augmenttraj}. 
By systematically designing, implementing, and evaluating selection strategies based on outlierness, diversity, representativeness, and uncertainty of core statistical properties, this work investigates the fundamental question of \textit{which} trajectories are most valuable for augmentation. This thesis provides an empirical comparison of these intelligent selection strategies with the aim of developing a practical, scalable framework for improving trajectory classification.

\subsection{Summary}
\label{sec:rw_summary}

The related work reveals that, while trajectory data augmentation is a new field, early efforts have demonstrated its potential. 
Many methods have focused on point perturbation~\cite{haranwala2023data, haranwala2023augmenttraj}, while related domains offer insight into generative approaches~\cite{agrawal2023trajaug}, validity preservation~\cite{antotsiou2021adversarial}, context-specific augmentation~\cite{mirkhani2024augmenting}, and intelligent sample selection~\cite{lin2023selectaugment}.

This thesis builds on the previously identified need for more systematic and intelligent trajectory selection for augmentation, with the goal of developing and evaluating strategies to improve trajectory classification performance by selecting which data to transform.

\newpage
\section{Method}
\label{sec:method}

This section describes the methodology used to investigate systematic trajectory selection strategies for data augmentation. 
The method is structured around developing, implementing, and evaluating four distinct selection strategies that go beyond the random approach established in the foundational work of Haranwala et al.~\cite{haranwala2023data}. 
The methodology ensures that the findings are credible, replicable and provide meaningful insight into the effectiveness of different trajectory selection approaches.

\subsection{Research Approach}
\label{subsec:research_approach}

% This study follows an experimental research approach, designed to evaluate and compare the performance of systematic trajectory selection strategies against random selection baselines. 
% The research methodology is quantitative, employing controlled experiments across multiple datasets and machine learning models to assess the impact of different selection strategies on trajectory classification performance. 

% The experimental design follows a comparative evaluation framework where multiple systematic selection strategies are implemented and tested, performance is measured using standardized evaluation metrics, results are compared against established baselines, and statistical significance is assessed across multiple experimental runs. 
% The choice of an experimental research design is justified by the need to provide empirical evidence on the effectiveness of systematic selection strategies, moving beyond theoretical propositions to practical validation of their utility in real-world trajectory classification scenarios.

This study adopts a quantitative experimental research approach to evaluate and compare systematic trajectory selection strategies with random baselines. 
Controlled experiments are conducted on multiple datasets and machine learning models, with performance assessed using standardized evaluation metrics and statistical significance tested on multiple runs. 
This experimental design provides empirical evidence of the effectiveness of systematic selection strategies in practical trajectory classification scenarios.

\begin{figure}[h] % or [ht]
    \centering
    \includegraphics[width=1\linewidth]{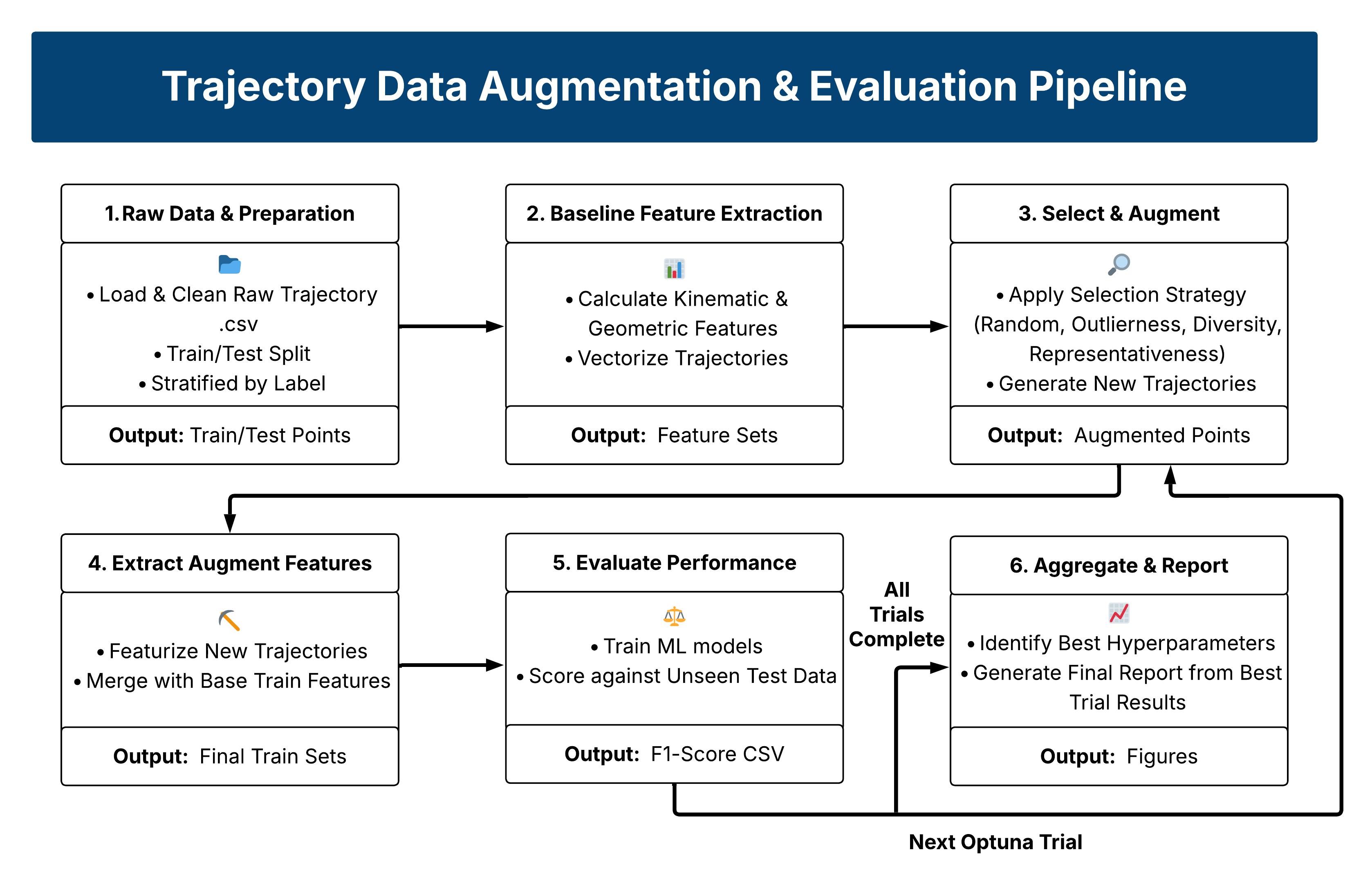}
    \caption{Trajectory Data Augmentation \& Evaluation Pipeline}
    \label{fig:pipeline}
\end{figure}

In this chapter, we start in Section \ref{subsec:selection_strategies} by detailing how trajectory selection strategies were mathematically defined. In Section \ref{subsec:experimental_framework}, we provide a comprehensive breakdown of the entire experimental pipeline, which is illustrated in Figure \ref{fig:pipeline}.

\noindent The pipeline is organized into six functional stages designed to ensure a rigorous and reproducible comparison between the selection methods. It begins with data preparation, where raw trajectory records are cleaned and partitioned into stratified training and testing sets. The second stage involves the extraction of baseline features to transform raw geographic coordinates into vectorized geometric and kinematic descriptors. In the third stage, the core experimental variable is introduced by applying systematic selection strategies to identify specific trajectories for augmentation and generating synthetic variants. Feature extraction is then performed on these new samples and merged with the original data to form the final training sets. The fifth stage consists of training machine learning models and scoring them against unseen test data using the weighted F1 score. Finally, the results of all all experimental runs are aggregated to produce the statistical reports and geometric visualizations used for analysis.

\subsection{Systematic Trajectory Selection Strategies}
\label{subsec:selection_strategies}

Building on the geographical noise techniques introduced by Haranwala et al.~\cite{haranwala2023data}, this work develops four systematic trajectory selection strategies. Each strategy implements a different theoretical approach to identifying which trajectories should be selected for augmentation. For a given class containing a set of trajectories $T$, each strategy aims to select a subset $S \subset T$ of size $|S| = p \cdot |T|$, where $p$ is the selection proportion.

\subsubsection{Random Selection}
\label{subsubsec:random_selection}

The random selection strategy serves as the simplest augmentation approach, replicating the original method used by Haranwala et al.~\cite{haranwala2023data}. Random selection chooses trajectories for augmentation using a uniform probability distribution, ensuring that each trajectory has an equal chance of being selected regardless of its characteristics. This strategy provides a baseline for comparing systematic selection approaches against unsystematic augmentation. \\

\textbf{Formalization:}
For a given class containing a set of trajectories $T$, the probability $P(t)$ of selecting any given trajectory $t$ of that class is uniform:
\begin{equation}
    P(t) = \frac{1}{|T|} \quad \forall t \in T
\end{equation}
Trajectories are sampled without replacement from each class until the target proportion $p$ is reached for that class.

The strategy randomly selects a specified proportion of trajectories from each class to maintain the balance of the class during augmentation. The selection process uses a fixed random seed to ensure reproducibility between experimental runs. Random selection serves as an important comparison point to demonstrate whether systematic approaches provide meaningful improvements over naive trajectory selection for augmentation.

\subsubsection{Outlierness-Based Selection}
\label{subsubsec:outlierness_selection}

The outlier-based selection strategy identifies trajectories that deviate significantly from typical patterns within their class. 
The underlying hypothesis is that augmenting outlier trajectories can help models better learn decision boundaries and improve the robustness to edge cases that may be underrepresented in the original dataset. 

This strategy employs the Distance-Based Outlier Score (DBOS) algorithm to quantify trajectory outlierness \cite{knorr1997_unified_approach_for_mining_outliers}. 
For each class, trajectory features are extracted. To establish a data-driven radius parameter $d$ for DBOS, the average pairwise Euclidean distance within the class's feature space is calculated. For computational feasibility on larger classes, this distance is estimated using a random sample ($N=1000$) of the population, avoiding the quadratic complexity of a full pairwise distance matrix while maintaining statistical validity. This adaptive radius is then used in the DBOS algorithm's ball-tree structure for efficient neighbor search, creating a neighborhood definition that adjusts to the specific data distribution of each class. \\

\textbf{Formalization:}
For each trajectory $t$ with feature vector $F(t)$ in a class set $T$, its neighborhood $N_d(t)$ is the set of other trajectories within radius $d$:
\begin{equation}
    N_d(t) = \{t' \in T \mid \text{distance}(F(t), F(t')) \le d\}
\end{equation}
The raw outlier score, $OS_{\text{raw}}(t)$, is defined by the inverted neighborhood count, where trajectories with fewer neighbors are assigned higher scores:
\begin{equation}
    OS_{\text{raw}}(t) = \max_{t' \in T} |N_d(t')| - |N_d(t)|
\end{equation}
This score is then scaled to a final value $OS(t)$ between 0 and 1 using min-max normalization to facilitate the ranking. The trajectory is ranked by $OS(t)$ in descending order, and the top proportion $p$ is selected for augmentation.

The DBOS algorithm's ball-tree structure allows for efficient neighbor searches to calculate outlier scores based on the number of neighboring trajectories within the specified radius. 
Trajectories with fewer neighbors receive higher outlier scores, indicating their deviation from typical class patterns. The trajectories are subsequently ranked by their outlier scores, and the highest-ranked trajectories are selected for augmentation. 
This approach ensures that the most unusual or edge-case trajectories within each class are prioritized for augmentation, potentially improving model robustness to diverse trajectory patterns.

\subsubsection{Diversity Maximization Selection}
\label{subsubsec:diversity_selection}

The diversity maximization strategy aims to select trajectories that collectively represent the maximum diversity within each class. This approach is based on the hypothesis that augmenting diverse trajectories will create a more comprehensive representation of the class space, leading to improved model generalization capabilities. The strategy uses k-means clustering to identify diverse trajectory groups within each class. \\ 

\textbf{Formalization:}
For each class, the set of trajectory feature vectors is partitioned into $k$ clusters, $C = \{C_1, C_2, \dots, C_k\}$, by minimizing the within-cluster sum of squares (WCSS):
\begin{equation}
    \arg\min_{C} \sum_{j=1}^{k} \sum_{F(t) \in C_j} \|F(t) - \mu_j\|^2
\end{equation}
where $\mu_j$ is the centroid of the cluster $C_j$. The selection is then performed by sampling proportionally from each cluster.

For each class, k-means clustering is applied to the trajectory feature space. The number of clusters was set to $k = 5$, a heuristic chosen to achieve a meaningful partitioning of the feature space; this value was adaptively reduced for any class with fewer than five trajectories. We acknowledge that using a fixed $k=5$ across domains as diverse as maritime traffic and animal movement is an arbitrary heuristic and a limitation of this study. While dynamically determining the optimal $k$ (e.g., via Silhouette scores or the elbow method) for each class would provide a more scientifically rigorous mapping of the dataset's variety, a fixed $k$ was chosen strictly as a computational trade-off to keep the hyperparameter optimization loop tractable.

Representatives are selected from each cluster, with selection proportional to cluster size to ensure that larger, more representative groups contribute more trajectories to the augmentation set. For computational efficiency with large datasets, the MiniBatchKMeans algorithm is employed instead of standard k-means clustering.

\subsubsection{Representativeness-Based Selection}
\label{subsubsec:representativeness_selection}

The representativeness-based selection strategy identifies trajectories that are most similar to the overall distribution of their class. 
The hypothesis underlying this approach is that augmenting representative trajectories will strengthen the model's understanding of typical class patterns, reinforcing the most common and characteristic behaviors within each class. 

This strategy uses statistical similarity measures to identify representative trajectories. 
For each class, z-scores are calculated for all trajectory features, normalizing the feature values relative to the class mean and standard deviation. 
A representativeness score is computed as the sum of absolute z-scores for each trajectory, where lower scores indicate trajectories that are closer to the class centroid in the normalized feature space. \\

\textbf{Formalization:}
For each feature $i$ in the feature vector $F(t)$ of a trajectory $t$, a z-score is calculated:
\begin{equation}
    z_i(t) = \frac{F_i(t) - \mu_i}{\sigma_i}
\end{equation}
where $\mu_i$ and $\sigma_i$ are the mean and standard deviation of feature $i$ across all trajectories in the class. The representativeness score $RS(t)$ is the L1 norm of the z-score vector, measuring the trajectory's total deviation from the class centroid:
\begin{equation}
    RS(t) = \sum_{i=1}^{m} |z_i(t)|
\end{equation}
where $m$ is the number of features. The trajectory with the lowest $RS(t)$ is considered the most typical and is selected.

Trajectories with lower representativeness scores are considered more representative of their class and are prioritized for selection. For robustness, all missing or non-finite feature values were consistently imputed with zero prior to score calculation. 
This approach selects trajectories that closely match the central tendency of their class, potentially reinforcing the model's ability to recognize and classify typical patterns.

\subsubsection{Uncertainty-Based Selection}

This strategy selects the trajectories that the model currently finds most confusing or ambiguous. The underlying hypothesis is that augmenting these \enquote{hard-to-classify} examples forces the model to refine its decision boundary in regions of high uncertainty, similar to the techniques used in active learning.  \\

\textbf{Formalization:} A \enquote{scout} logistic regression model is trained on the current dataset. For each trajectory $t$, the model predicts a probability distribution over all classes. The uncertainty score $U(t)$ is computed using entropy:
\begin{equation}
U(t) = - \sum_{c \in C} P(y = c \mid t)\,\log P(y = c \mid t),
\end{equation}
where $P(y = c \mid t)$ denotes the predicted probability that trajectory $t$ belongs to class $c$.

Trajectories with higher entropy scores correspond to cases in which the model exhibits a lower confidence in the classification. These trajectories are ranked in descending order of uncertainty, and the top proportion $p$ is selected for augmentation. To preserve class balance, the ranking and selection process is performed independently within each class.

\subsection{Experimental Framework}
\label{subsec:experimental_framework}

% The experimental framework integrates trajectory feature extraction, systematic selection, geographical perturbation, and performance evaluation into a comprehensive pipeline. 
% This framework ensures consistent evaluation across all selection strategies while maintaining the integrity of the augmentation process established in prior work.

\subsubsection{Dataset Preparation}
\label{subsubsec:dataset_preparation}

To ensure the generalizability of our findings, this study evaluates selection strategies across four diverse trajectory datasets. The datasets were selected to represent a range of movement patterns, domains, and spatial and temporal scales, including animal behavior, human mobility, and maritime traffic. Each dataset is split into 80\% training and 20\% testing sets using stratified sampling to preserve class distributions. To ensure that stratification is feasible, a preprocessing step removes classes with fewer than two samples.

In addition, several data quality checks are applied. Trajectories with zero variance in latitude and longitude are treated as stationary artifacts and excluded to avoid errors in kinematic feature computation. Trajectory identifiers (TIDs) are also explicitly cast to string types to prevent integer overflow and unintended formatting issues. This entire procedure is repeated on 20 random seeds to ensure statistical robustness. \\

The final datasets selected for this research are the following:

\begin{itemize}
    
    \item \textbf{Foxes:} This dataset tracks the movement of 66 adult arctic foxes (\textit{Vulpes lagopus}) on Bylot Island, Nunavut, Canada. Originally sourced from a six year study using Argos satellite telemetry, the data capture how these predators navigate a fluctuating tundra environment \cite{lai2017movement}. The dataset comprises 229,657 data points and is utilized here for gender classification. With an average trajectory length of 3,480 points, it represents a complex record of biological movement patterns.

    \item \textbf{Starkey:} This dataset contains movement records for 253 ungulates, including elk and deer, within the Starkey Experimental Forest and Range in Oregon, United States \cite{wisdom2005starkey}. It consists of 287,136 data points categorized into three behavioral classes. The trajectories in this dataset are relatively short, averaging 1,135 points per record. It serves as a benchmark for classifying animal behavior and habitat use in a controlled environment.

    \item \textbf{Car Traffic:} This dataset captures the movement of 125 vehicles in Brno, Czech Republic, and is sourced from the example collections of the PTRAIL library \cite{haidri2022ptrail}. It contains exactly 44,905 data points categorized into two classes: car and bus. The trajectories are characterized by a short average length of 359 points. This dataset is included to evaluate how augmentation selection strategies perform in high velocity urban mobility contexts.

    \item \textbf{AIS Subset:} The original maritime dataset provided for this research, which contains millions of vessel trajectories, was too large for the computational scope of the project. Therefore, a specialized subset was created. This data is derived from the Automatic Identification System (AIS), a tracking system used ship-tracking system that broadcasts a ship's speed. To create a manageable and balanced multi-class dataset, the original ais.csv file was processed programmatically. The script iterated through the large file in chunks, filtering for trajectories belonging to four specific vessel types: Fishing (label 30), Tug (label 52), Passenger Ship (label 60), and Cargo Ship (label 70). The resulting ais\_subset.csv file contains 694 unique trajectories and a total of 2,660,535 data points, providing a challenging and realistic scenario for maritime trajectory classification.
\end{itemize}

\noindent With these datasets prepared, the next step is a systematic feature-extraction process to create a robust numerical representation for each trajectory.

\subsubsection{Feature Extraction}
\label{subsubsec:feature_extraction}
To create a robust representation of trajectory behavior, the framework employs a systematic process to derive features capturing geometric, kinematic, and statistical properties. 
These features serve a dual purpose: they inform the selection algorithms and provide input tonal machine learning classifiers. The implementation is optimized for performance, using vectorized calculations to efficiently process large datasets.

\begin{enumerate}
    \item \textit{Distance-based Geometry Features:} The shape of each trajectory is captured through a multi-level sinuosity analysis. Trajectories are partitioned into segments at five distinct levels of granularity (1-5 segments). 
    For each segment, the ratio of the geodesic distance (straight-line) to the cumulative path distance is calculated. To optimize this process, the cumulative path distance is computed from a precomputed set of inter-point segments using fast, vectorized ellipsoidal geodesic calculations (WGS84) via the pyproj library. 
    This process quantifies the straightness of the trajectory versus the tortuosity on multiple scales of analysis, yielding 15 geometric features. 
    To overcome the computational overhead of iterative fractal summation on large datasets, these specific calculations were optimized using Numba Just-In-Time (JIT) compilation, achieving C-level execution speeds within the Python environment.

    \item \textit{Kinematic and Statistical Features:} Movement dynamics is characterized by three core kinematic measures: speed, acceleration, and turning angle. The implementation leverages vectorized NumPy operations to efficiently compute these measures across entire trajectories at once, avoiding slow, iterative calculations. To go beyond simple averages, a comprehensive statistical profile is then computed for each of these three measures. This includes 19 statistical metrics, such as mean, standard deviation, skewness, kurtosis, median absolute deviation, and a range of quantiles (e.g., 5th, 25th, 75th, 95th). This comprehensive profiling captures the full distribution of movement behaviors, providing a nuanced view of a trajectory's dynamics.
\end{enumerate}
\noindent The complete feature extraction process generates a 72-dimensional feature vector for each trajectory (15 from geometry + 3 kinematic measures $\times$ 19 statistical features), creating a rich, multi-faceted representation suitable for complex behavioral analysis.

\subsubsection{Augmentation Process}
\label{subsubsec:augmentation_process}

The augmentation process employs the \enquote{on-circle} noise strategy from Haranwala et al.~\cite{haranwala2023data} to apply realistic geographical perturbations to selected trajectories. 
The methodology is defined by its core perturbation mechanism, the hyper-parameters that govern it, and a rigorous seed management protocol that ensures all experimental results are fully reproducible and statistically sound. 
The systematic method for selecting optimal values for these hyper-parameters is detailed in Section~\ref{subsubsec:parameter_justification}.

\begin{enumerate}
    \item \textit{Reproducibility and Seed Management:} To ensure the scientific validity and complete reproducibility of the experiments, a meticulous seed management protocol is enforced. The entire experimental framework uses 20 master seeds, derived from consecutive digits of $\pi$ (1415, 9265, 3589, etc.). Each master seed controls one complete experimental run, dictating the randomness for dataset splitting, trajectory selection, and the augmentation operations within that run. Crucially, for any given augmentation operation, a unique and deterministic seed is calculated from the master seed, the augmentation number (from 1 to $n$), and the trajectory's unique ID. This multi-level seeding strategy is fundamental, as it ensures that when comparing the effects of different selection strategies, the underlying randomness of the perturbation applied to a specific trajectory remains constant, thereby robustly isolating the impact of the selection method itself.

    \item \textit{Geographical Perturbation Mechanism:} The on-circle strategy modifies a trajectory by relocating a subset of its \textit{interior} points, thereby preserving the original start and end points to maintain the trajectory's overall structural integrity. For each point chosen for perturbation, it is moved to a new random location on the circumference of a circle centered at its original position. A critical feature of this implementation is the dynamic scaling of the radius of the circle. The radius is not a fixed value, but is set instead at 50\% of the distance to the \textit{next} point in the trajectory. This adaptive approach is fundamental because it ensures that perturbations are context-aware. In dense, slow-moving segments where the points are close together, the perturbation radius is small, preserving the fine-grained structure of the movement. In contrast, in sparse, fast-moving segments, the radius is larger, introducing more significant (but still plausible) geographical noise. This prevents the generation of unrealistic and jerky movements and ensures that the augmented data remain semantically coherent with the original movement patterns.

    \item \textit{Governing Hyperparameters:} The intensity and scale of the augmentation process are controlled by a set of three key hyperparameters. These parameters serve as the primary variables in the experimental design, enabling a systematic exploration of the impact of the augmentation strategy, namely:
    \begin{itemize}
        \item The \textit{Trajectory Selection Proportion ($p$)}, which dictates what fraction of trajectories to augment.
        \item The \textit{Number of Augmentations ($n$)}, which defines how many new versions are created for each selected trajectory.
        \item The \textit{Point Modification Proportion ($pp$)}, which determines the proportion of interior points within a selected trajectory that are to be perturbed.
    \end{itemize}
\end{enumerate}

\noindent Each newly generated trajectory is assigned a unique identifier and metadata that trace it back to its parent trajectory. The final augmented dataset for training is a union of the original trajectories and their newly generated counterparts, enriching the training data while preserving the original class distribution.

\subsubsection{Model Training and Evaluation}
\label{subsubsec:model_training_evaluation}

To strictly evaluate the impact of the augmentation strategies, a diverse suite of four machine learning models was employed: Random Forest, XGBoost, a Multi-Layer Perceptron (MLP), and Logistic Regression. This selection spans multiple algorithmic families to ensure the generalizability of the findings.

A primary challenge in this experimental design is the vast search space created by combining model hyper-parameters with augmentation hyper-parameters. A simultaneous, exhaustive optimization of both was computationally infeasible. Such an approach would lead to a combinatorial explosion, as the 27 augmentation configurations (3 trajectory selection proportions $\times$ 3 augmentation counts $\times$ 3 point modification proportions) would need to be cross-multiplied with the extensive hyperparameter grids for each of the four models. For a single model such as XGBoost, this would have resulted in more than 11,000 combinations to test in each experimental run.

To overcome this computational barrier, the parameter search spaces were decoupled. 
This led to the adoption of a systematic two-phase tuning process that is not only computationally tractable but also methodologically more robust. 
This \enquote{baseline-first} approach is designed to isolate the effect of data augmentation from the effect of model hyperparameter tuning.

\begin{enumerate}
    \item \textit{Baseline Model Tuning:} For each dataset and each of the 20 master seeds, the models are first tuned exclusively on the original, non-augmented 80\% training data. This is performed using a 3-fold cross-validated Grid Search (GridSearchCV) to find the optimal hyperparameter configuration for each model under baseline conditions. These baseline parameters that perform best are then serialized and cached on disk. This strictly enforces that subsequent runs on augmented data use the exact same model configuration, isolating the data quality as the only variable.
    
    \item \textit{Augmentation Evaluation:} When evaluating a given augmentation strategy (e.g., Outlierness with a specific set of Optuna-derived hyperparameters), the models are retrained on the newly augmented dataset. Crucially, they are trained using the optimal baseline parameters discovered and cached in the first phase. 
\end{enumerate}

\noindent This protocol ensures that any observed performance difference between the baseline and an augmented version can be attributed directly to the quality of the augmented training data, rather than to coincidental benefits from simultaneous model re-tuning.

The selected models and their corresponding hyperparameter search spaces for the initial \textit{GridSearchCV} tuning are as follows: \\

\textit{Random Forest:} An ensemble method using multiple decision trees.

\begin{itemize}
    \item n\_estimators: [100, 300, 500]
    \item max\_depth: [10, 20, 40, None]
    \item min\_samples\_leaf: [1, 2, 5]
    \item max\_features: ['sqrt', 'log2']
    \item criterion: ['gini', 'entropy']
\end{itemize}

\textit{XGBoost:} A gradient boosting framework using decision trees.
\begin{itemize}
    \item n\_estimators: [100, 300, 500]
    \item max\_depth: [3, 5, 8]
    \item learning\_rate: [0.02, 0.05, 0.1, 0.2]
    \item subsample: [0.8, 1.0]
    \item colsample\_bytree: [0.8, 1.0]
    \item gamma: [0, 0.1, 0.5]
\end{itemize}

\textit{MLP:} A feedforward artificial neural network (Multi-Layer Perceptron).
\begin{itemize}
    \item classifier\_\_hidden\_layer\_sizes: [(50, 50), (100,), (200,), (100, 50), (100, 100), (100, 50, 50), (100, 100, 50), (100, 100, 100)]
    \item classifier\_\_alpha: [0.0001, 0.001, 0.01]
    \item classifier\_\_activation: ['relu', 'tanh']
    \item classifier\_\_learning\_rate\_init: [0.001, 0.01]
\end{itemize}

\textit{Logistic Regression:} A linear model for binary/multiclass classification.
\begin{itemize}
    \item classifier\_\_C: [0.01, 0.1, 1.0, 10.0, 100.0]
    \item classifier\_\_penalty: ['l1', 'l2']
    \item classifier\_\_solver: ['saga']
\end{itemize}

\noindent Performance is consistently measured in the same 20\% holding test set for all experiments. The evaluation metric used is the weighted F1-score, which is chosen to properly account for potential class imbalances in the datasets. The final reported results are the average scores across all 20 master seeds, ensuring robust and statistically sound conclusions.

\subsubsection{Parameter Selection and Justification}
\label{subsubsec:parameter_justification}

A core contribution of this thesis is to move beyond a \enquote{one-size-fits-all} approach to augmentation. Preliminary research and the diverse nature of trajectory datasets suggest that a single fixed set of augmentation parameters is unlikely to be optimal across all contexts. For example, a data-scarce dataset may benefit from a more aggressive augmentation strategy than a large, dense dataset.

Therefore, this work departs significantly from prior methods that rely on predefined parameters. Instead, it implements a systematic hyperparameter optimization framework to empirically determine the most effective augmentation configurations on a per-dataset basis. This ensures that the evaluation is not based on an arbitrary configuration but rather on the strongest possible application of each augmentation strategy for a given dataset.

To manage this search, the \textit{Optuna} hyperparameter optimization library is integrated into the experimental pipeline. For each dataset, an optimization \enquote{study} is conducted with the goal of finding the combination of augmentation parameters that maximizes model performance. The metric to be maximized is the average weighted F1-score across all four machine learning models and all 20 master seeds for a given parameter set.

The optimization process explores the three governing hyperparameters defined in Section~\ref{subsubsec:augmentation_process}. A discrete search space was defined for each, balancing a wide range of intensities of augmentation with computational feasibility. To ensure that the true optimum within this space was always found, the study was configured to use Optuna's \textit{GridSampler}. This strategy performs an exhaustive grid search on all possible combinations rather than a random or probabilistic search. \\

The search space for the 27 unique augmentation configurations is as follows:
\begin{itemize}
    \item \textit{Trajectory Selection Proportion ($p$):} [0.1, 0.2, 0.4]
    \item \textit{Number of Augmentations ($n$):} [1, 3, 5]
    \item \textit{Point Modification Proportion ($pp$):} [0.1, 0.2, 0.4]
\end{itemize}

\noindent While the high-level augmentation strategy was optimized, the low-level perturbation mechanism parameter remained fixed, as its adaptive nature makes it broadly applicable.
\begin{itemize}
    \item \textit{Circle Radius (50\% of the distance to the next point):} This parameter was kept constant throughout all experiments. To clarify, this does not mean the spatial radius was a static global value. Because it is calculated as a percentage of the distance between two specific consecutive points, the physical radius scales dynamically with the object's velocity. For a fast-moving car where points are miles apart, the perturbation radius expands to miles; for a slow-moving fox, the radius shrinks to meters. Its dynamic, data-driven scaling provides a highly adaptive, localized perturbation. For each point chosen for modification, the radius is recalculated. This ensures that perturbations are context-aware: smaller in dense, slow-moving segments to preserve fine-grained structure, and larger in sparse, fast-moving segments to introduce meaningful variation. This approach has been validated in previous research~\cite{haranwala2023data} and effectively generates realistic geographical noise regardless of the dataset's domain, from local animal movements (Foxes) to urban mobility (Car Traffic).
\end{itemize}

\noindent This optimization-centric approach enables a data-driven, per-dataset determination of the optimal augmentation strategy. The final results, presented in Section~\ref{sec:results_analysis}, are therefore based not on a fixed assumption but on an empirical comparison of the most effective augmentation configurations for each specific data context, ensuring a fair and robust evaluation of the selection strategies themselves.

\subsection{Implementation Details}
\label{subsec:implementation}

The complete methodology is implemented in Python, using several specialized libraries. Core dependencies include \textit{pandas} and \textit{numpy} for data manipulation; \textit{scikit-learn} for machine learning algorithms and preprocessing; \textit{xgboost} for the gradient boosting implementation; and \textit{Optuna} for the hyperparameter optimization framework. A critical technical enhancement involved the use of the \textit{pyproj} library for geographical calculations. Computational bottlenecks were addressed by replacing iterative distance metrics with vectorized operations using \textit{pyproj}. By mapping the WGS84 ellipsoid calculations directly to the underlying C arrays, the feature extraction and augmentation phases achieved a speedup of several orders of magnitude compared to standard iterative approaches, allowing efficient processing of millions of trajectory points.

Computational considerations are further addressed through several optimization strategies. A multi-level parallel processing strategy is employed. At the experiment level, multiprocessing distributes independent dataset seeds across available CPU cores, while at the model level, scikit-learn utilizes threaded parallelism for \textit{GridSearch} operations. Explicit memory management (garbage collection) was implemented to prevent RAM saturation during these concurrent high-load operations. Additionally, I/O overhead was minimized using the LZ4-compressed Feather binary format for intermediate data storage. Unlike CSVs, Feather files preserve native Pandas data types, and the LZ4 compression significantly reduces disk I/O bottlenecks and storage footprint without compromising the high-speed read/write throughput required for repeatedly loading large trajectory datasets across multiple experimental seeds. To ensure transparency and reproducibility, the complete source code for this framework has been archived and made publicly available~\cite{nordling_2025_code}. 

The pipeline also features a multi-level caching system to maximize computational efficiency. At a low level, it avoids redundant feature extraction by checking for existing feature files before processing raw data. At a higher level, the Optuna study incorporates a robust caching mechanism that saves the final F1-score for each completed trial. This prevents the costly re-execution of entire experimental runs if the study is interrupted and resumed, or if the same parameters are evaluated more than once.

The implementation follows best practices in software engineering, including modular design for maintainability, comprehensive error handling, and extensive logging. To ensure the resilience of the pipeline, a robust fallback mechanism was implemented. In cases where systematic strategies encountered mathematical instabilities, such as classes with zero feature variance that prevented a valid calculation of the outlier score, the system automatically defaulted to random selection for that specific class. This prevented experimental failure while preserving the maximum amount of usable data. This modular architecture enables easy extension with additional selection strategies, supporting future research directions.

\subsection{Ethical Considerations}
\label{subsec:ethics}

This research involves the analysis of trajectory datasets that may contain location information with potential privacy implications. Several ethical considerations are carefully addressed to ensure responsible research conduct. All datasets used in this study are publicly available and have been previously published in academic contexts with appropriate permissions and ethical clearances. During research, no personally identifiable information is processed or stored and all analysis is conducted on aggregated or anonymized trajectory data. 

The data augmentation techniques are designed to preserve privacy by introducing geographical noise that obscures exact locations while preserving overall movement patterns. The results are reported in aggregate form without exposing individual trajectory patterns or specific location information. The research contributes positively to the field by enabling the development of better machine learning models with limited data, potentially reducing the need to collect additional sensitive trajectory information from individuals.

\subsection{Summary}
\label{subsec:summary}

This section has described a comprehensive methodology for investigating systematic trajectory selection strategies in data augmentation. The approach combines theoretical foundations from outlier detection, diversity maximization, and representativeness analysis with practical implementation in a robust, optimization-driven experimental framework. The methodology enables a rigorous, data-driven comparison of selection strategies across multiple datasets and models, providing empirical evidence for their effectiveness in improving trajectory classification performance.

The systematic approach ensures that the results are reliable, valid and generalizable across different trajectory domains. The next section will present the detailed results of applying this methodology, demonstrating the comparative effectiveness of different selection strategies and their impact on the performance of the machine learning model across the four evaluation datasets.

\newpage
\section{Results and Analysis}
\label{sec:results_analysis}

This section presents the findings of the systematic evaluation of the trajectory selection strategy. The performance of each selection strategy was evaluated across the four designated datasets (AIS Subset, Car Traffic, Foxes, and Starkey) using four machine learning models (Logistic Regression, MLP, Random Forest, and XGBoost). The primary metric for evaluation is the weighted F1-score, averaged across twenty different random seeds to ensure statistical robustness and mitigate the effects of random initialization.

\begin{table}[ht]
    \centering
\caption{Performance comparison using the optimal augmentation hyperparameters identified for each dataset. Each score represents the weighted F1-score, averaged over all seeds, for that specific model and strategy. Scores higher than the baseline are in \textbf{bold}.\\ 
    For each dataset, the selected parameters are shown in parentheses:\\ 
    \textit{{Sel} = proportion of trajectories selected}\\ 
    \textit{{Augs} = number of augmentations per trajectory}\\ 
    \textit{{Pts} = proportion of points modified.}}
    \label{tab:final_results}

    \sisetup{table-format=2.2, table-space-text-post={\%}}

    \resizebox{\linewidth}{!}{
    \begin{tabular}{l S S S S S S}
        \toprule
        \textbf{Model} & {\textbf{Baseline}} & {\textbf{Diversity}} & {\textbf{Outlierness}} & {\textbf{Random}} & {\textbf{Representativeness}} & {\textbf{Uncertainty}} \\
        \midrule
        \multicolumn{7}{l}{\textit{\textbf{Ais Subset Dataset}} \hfill \footnotesize (Sel: 20\%, Augs: 1, Pts: 10\%)} \\
        LogisticReg     & 68.54\% & \textbf{68.55\%} & \textbf{68.61\%} & 67.99\% & \textbf{68.95\%} & \textbf{68.63\%} \\
        MLP             & 68.14\% & \textbf{68.26\%} & \textbf{68.77\%} & \textbf{68.53\%} & \textbf{69.57\%} & \textbf{69.75\%} \\
        RandomForest    & 76.48\% & 76.45\% & \textbf{76.98\%} & \textbf{76.50\%} & \textbf{76.52\%} & 76.33\% \\
        XGBoost         & 74.55\% & \textbf{74.73\%} & \textbf{75.09\%} & \textbf{74.69\%} & \textbf{74.99\%} & \textbf{74.96\%} \\
        \midrule
        \multicolumn{7}{l}{\textit{\textbf{Car Traffic Dataset}} \hfill \footnotesize (Sel: 20\%, Augs: 3, Pts: 10\%)} \\
        LogisticReg     & 84.69\% & \textbf{85.08\%} & \textbf{84.87\%} & 84.44\% & \textbf{85.10\%} & 83.41\% \\
        MLP             & 84.39\% & \textbf{85.64\%} & \textbf{84.53\%} & 83.87\% & \textbf{86.05\%} & \textbf{84.99\%} \\
        RandomForest    & 82.28\% & \textbf{82.96\%} & 82.18\% & \textbf{82.57\%} & 82.18\% & \textbf{82.38\%} \\
        XGBoost         & 85.89\% & \textbf{86.55\%} & \textbf{87.01\%} & 84.95\% & 84.35\% & 83.92\% \\
        \midrule
        \multicolumn{7}{l}{\textit{\textbf{Foxes Dataset}} \hfill \footnotesize (Sel: 40\%, Augs: 5, Pts: 10\%)} \\
        LogisticReg     & 42.25\% & \textbf{46.83\%} & \textbf{49.60\%} & \textbf{47.59\%} & \textbf{49.07\%} & \textbf{47.08\%} \\
        MLP             & 46.78\% & 41.64\% & \textbf{50.54\%} & \textbf{47.73\%} & \textbf{48.19\%} & \textbf{53.69\%} \\
        RandomForest    & 43.75\% & \textbf{45.11\%} & \textbf{47.44\%} & \textbf{47.69\%} & \textbf{51.35\%} & \textbf{52.01\%} \\
        XGBoost         & 50.94\% & 48.83\% & \textbf{51.03\%} & \textbf{53.49\%} & \textbf{51.74\%} & 50.60\% \\
        \midrule
        \multicolumn{7}{l}{\textit{\textbf{Starkey Dataset}} \hfill \footnotesize (Sel: 20\%, Augs: 3, Pts: 10\%)} \\
        LogisticReg     & 88.32\% & \textbf{88.61\%} & \textbf{88.75\%} & \textbf{88.84\%} & \textbf{88.34\%} & 88.13\% \\
        MLP             & 86.22\% & \textbf{88.20\%} & \textbf{88.22\%} & \textbf{88.30\%} & \textbf{87.66\%} & \textbf{89.53\%} \\
        RandomForest    & 92.00\% & 91.39\% & 90.38\% & 91.64\% & 90.65\% & 91.38\% \\
        XGBoost         & 90.37\% & \textbf{90.67\%} & 88.96\% & 90.17\% & 89.45\% & \textbf{90.75\%} \\
        \bottomrule
    \end{tabular}
    }
\end{table}

\subsection{Overall Performance Comparison}
\label{sec:overallresults}
The central outcome of the experimental pipeline is a comparison of the performance of the baseline model with the performance achieved using each of the five systematic augmentation strategies: Diversity, Outlierness, Random, Representativeness, and Uncertainty.

Table \ref{tab:final_results} presents a high-level summary of the optimal performance achieved for each dataset. The values represent the mean weighted F1-score in all 20 seeds. A critical component of these results is the dataset-specific optimization performed by the Optuna framework. As indicated in the table headers, the optimal augmentation configuration defined by the Selection Proportion ($Sel$), Number of Augmentations ($Augs$), and Point Modification Proportion ($Pts$) varied significantly between domains. This variance underscores that the intensity of augmentation must be tailored to the specific sparsity and noise characteristics of the underlying data.

The tabulated results indicate a distinct heterogeneity in the effectiveness of data augmentation in different data contexts. The \textit{Foxes} dataset, characterized by sparsity and complex animal movement, exhibited the most substantial response to augmentation. In this data-scarce environment, the baseline models struggled to generalize, achieving scores between 42\% and 50\%. However, the introduction of systematic augmentation resulted in transformative gains, with the Representativeness and Uncertainty strategies pushing performance up by more than 8 to 11 percentage points for the Random Forest model. This confirms the hypothesis that systematic augmentation acts as a powerful regularizer in high-variance, low-data regimes.

In contrast to the sparse Foxes dataset, the \textit{Car Traffic} and \textit{Starkey} datasets possessed high baseline F1-scores, ranging from 82\% to over 90\%. These domains proved far more resistant to augmentation. For the Car Traffic dataset, the gains were marginal or non-existent, illustrating a saturation point where the decision boundaries were already well-defined by the original data. In such high-quality data environments, additional synthetic data yield diminishing returns and, in some cases, risk introducing noise that degrades the classifier's precision.

A comparison of the selection strategies reveals a consistent trend where systematic approaches outperformed the naive Random baseline. Although random selection occasionally provided marginal benefits, it consistently trailed behind strategies such as Outlierness and Uncertainty. For instance, in the AIS Subset, the Random strategy failed to improve upon the baseline for the Logistic Regression model, whereas the Representativeness and Uncertainty strategies achieved notable gains. This validates the core premise of this thesis that the strategic selection of \textit{which} trajectories to augment is as critical as the augmentation mechanism itself.

Furthermore, the results highlight a strong interaction between the selection strategy and the model architecture. Non-linear models, particularly the Multi-Layer Perceptron (MLP), demonstrated a unique capacity to leverage the Uncertainty strategy. Although tree-based models such as Random Forest often plateaued on dense AIS and Starkey datasets, the MLP model achieved its highest scores when trained on data augmented via Uncertainty selection. This suggests that neural networks, which optimize non-convex loss functions, benefit significantly from active-learning-style data injection that targets ambiguous regions of the feature space, preventing convergence to suboptimal local minima.

\subsection{Statistical Analysis by Dataset}
To validate whether the observed improvements are statistically significant or simply artifacts of random seed variance, a rigorous paired t-test ($\alpha=0.05$) was conducted for each dataset. 

It is important to clarify the distinction between this analysis and the previous summary. Although Table \ref{tab:final_results} presented the absolute weighted F1-scores, the following analysis isolates the \textit{relative performance gain} (or loss) achieved by the optimal augmentation configuration compared to its specific seed-matched baseline. This paired comparison filters out the variance inherent in random initialization, providing a stricter test of whether the augmentation strategy truly contributes value beyond the baseline variance.

\subsubsection{AIS Subset}
The \textit{AIS Subset} represents a challenging maritime environment with high density and complex vessel interaction. As shown in Table \ref{tab:final_analysis_ais_subset}, the high baseline performance of the tree-based models made further improvements difficult. For Logistic Regression, Random Forest, and XGBoost, no strategy yielded statistically significant improvements ($p > 0.05$), and mean improvements were negligible or slightly negative.

However, a critical breakthrough with the Multi-Layer Perceptron is observed. The Uncertainty strategy achieved a statistically significant mean improvement of +1.61\% ($p = 0.0441$). This suggests that in a dense maritime dataset, while the typical trajectory of a vessel is easily learned, the neural network struggles primarily with ambiguous transition zones between classes. The Uncertainty strategy likely targets these ambiguous instances for augmentation, forcing the model to refine its decision boundary in the most confusing regions of the manifold.

% Auto-generated for ais_subset with 20 seeds.
\begin{table}[ht]
    \centering
    \caption{
        Statistical analysis of the performance improvement for each optimally-tuned 
        augmentation strategy compared to the baseline. This table presents the final, 
        rigorous comparison, in contrast to the exploratory overview in the previous table.
        An asterisk (*) denotes a statistically significant result (p < 0.05).
    }
    \label{tab:final_analysis_ais_subset}
    \begin{tabular}{{l l r r}}
        \toprule
        \textbf{Model} & \textbf{Strategy} & \textbf{Mean Improv. (\%)} & \textbf{p-value} \\
        \midrule
        \multirow{5}{*}{\textbf{LogisticRegression}} 
        & Diversity & +0.01 & 0.9868 \\
        & Outlierness & +0.07 & 0.8814 \\
        & Random & -0.09 & 0.8073 \\
        & Representativeness & +0.41 & 0.2176 \\
        & Uncertainty & +0.09 & 0.8161 \\
        \midrule
        \multirow{5}{*}{\textbf{MLP}} 
        & Diversity & +0.12 & 0.8949 \\
        & Outlierness & +0.63 & 0.4374 \\
        & Random & -0.08 & 0.9155 \\
        & Representativeness & +1.43 & 0.1444 \\
        & \textbf{Uncertainty} & \textbf{+1.61} & \textbf{0.0441*} \\
        \midrule
        \multirow{5}{*}{\textbf{RandomForest}} 
        & Diversity & -0.03 & 0.9308 \\
        & Outlierness & +0.50 & 0.0936 \\
        & Random & +0.05 & 0.8693 \\
        & Representativeness & +0.04 & 0.8040 \\
        & Uncertainty & -0.16 & 0.6194 \\
        \midrule
        \multirow{5}{*}{\textbf{XGBoost}} 
        & Diversity & +0.18 & 0.5423 \\
        & Outlierness & +0.54 & 0.1878 \\
        & Random & +0.48 & 0.1423 \\
        & Representativeness & +0.44 & 0.1335 \\
        & Uncertainty & +0.41 & 0.2969 \\
        \bottomrule
    \end{tabular}
\end{table}

\newpage

\subsubsection{Car Traffic}
The \textit{Car Traffic} dataset, representing high-velocity urban mobility, presented a scenario of high model saturation. As detailed in Table \ref{tab:final_analysis_car_traffic}, the baseline performance of all models was already robust, ranging between 82\% and 85\%.

Statistical analysis reveals that no augmentation strategy provided a statistically significant improvement for any model in this dataset. Although the MLP model saw a mean improvement of +1.25\% using the Diversity strategy and +1.66\% using Representativeness, the high p-values (0.2958 and 0.1684, respectively) indicate high variance between seeds, suggesting that these gains are not reliable. The resistance of this dataset to augmentation implies that the decision boundaries are already well-defined by the original data, or that the physical constraints of road networks make standard geometric perturbation less effective at generating informative synthetic samples.

% Auto-generated for car_traffic with 20 seeds.
\begin{table}[ht]
    \centering
    \caption{
        Statistical analysis of the performance improvement for each optimally-tuned 
        augmentation strategy compared to the baseline. This table presents the final, 
        rigorous comparison, in contrast to the exploratory overview in the previous table.
        An asterisk (*) denotes a statistically significant result (p < 0.05).
    }
    \label{tab:final_analysis_car_traffic}
    \begin{tabular}{{l l r r}}
        \toprule
        \textbf{Model} & \textbf{Strategy} & \textbf{Mean Improv. (\%)} & \textbf{p-value} \\
        \midrule
        \multirow{5}{*}{\textbf{LogisticRegression}} 
        & Diversity & +0.38 & 0.8107 \\
        & Outlierness & +0.18 & 0.8891 \\
        & Random & -1.05 & 0.2057 \\
        & Representativeness & +0.41 & 0.6423 \\
        & Uncertainty & -0.10 & 0.8932 \\
        \midrule
        \multirow{5}{*}{\textbf{MLP}} 
        & Diversity & +1.25 & 0.2958 \\
        & Outlierness & +0.14 & 0.8695 \\
        & Random & +1.06 & 0.4347 \\
        & Representativeness & +1.66 & 0.1684 \\
        & Uncertainty & -0.69 & 0.3838 \\
        \midrule
        \multirow{5}{*}{\textbf{RandomForest}} 
        & Diversity & +0.68 & 0.2515 \\
        & Outlierness & -0.10 & 0.5770 \\
        & Random & +0.10 & 0.3299 \\
        & Representativeness & -0.10 & 0.3299 \\
        & Uncertainty & +0.10 & 0.3299 \\
        \midrule
        \multirow{5}{*}{\textbf{XGBoost}} 
        & Diversity & +0.66 & 0.4829 \\
        & Outlierness & +1.12 & 0.2712 \\
        & Random & +0.18 & 0.8601 \\
        & Representativeness & -1.54 & 0.2323 \\
        & Uncertainty & -1.20 & 0.2469 \\
        \bottomrule
    \end{tabular}
\end{table}

\newpage

\subsubsection{Foxes}
The \textit{Foxes} dataset represents the data-scarce scenario, where the baseline models performed poorly. Table \ref{tab:final_analysis_foxes} shows that this domain benefited the most dramatically from systematic augmentation. Unlike previous datasets, nearly all systematic strategies delivered substantial improvements. For Logistic Regression, the Diversity, Outlierness and Representativeness strategies all yielded statistically significant gains, with Outlierness improving the score by more than 7\%.

The most profound impact was observed in the Random Forest model. Here, the Representativeness strategy achieved a remarkable mean improvement of +11.56\% ($p = 0.0004$), and the Uncertainty strategy improved performance by +8.26\% ($p = 0.0331$). This confirms that in sparse and noisy biological datasets, systematic augmentation acts as a powerful regularizer, synthesizing the necessary data volume that allows complex models to generalize rather than overfit. Notably, the naive Random selection often trailed behind systematic approaches or failed to achieve significance, validating the thesis hypothesis.

% Auto-generated for foxes with 20 seeds.
\begin{table}[ht]
    \centering
    \caption{
        Statistical analysis of the performance improvement for each optimally-tuned 
        augmentation strategy compared to the baseline. This table presents the final, 
        rigorous comparison, in contrast to the exploratory overview in the previous table.
        An asterisk (*) denotes a statistically significant result (p < 0.05).
    }
    \label{tab:final_analysis_foxes}
    \begin{tabular}{{l l r r}}
        \toprule
        \textbf{Model} & \textbf{Strategy} & \textbf{Mean Improv. (\%)} & \textbf{p-value} \\
        \midrule
        \multirow{5}{*}{\textbf{LogisticRegression}} 
        & \textbf{Diversity} & \textbf{+5.80} & \textbf{0.0454*} \\
        & \textbf{Outlierness} & \textbf{+7.18} & \textbf{0.0153*} \\
        & Random & +4.94 & 0.0691 \\
        & \textbf{Representativeness} & \textbf{+4.84} & \textbf{0.0456*} \\
        & Uncertainty & +4.82 & 0.2079 \\
        \midrule
        \multirow{5}{*}{\textbf{MLP}} 
        & Diversity & +0.87 & 0.7930 \\
        & Outlierness & +3.35 & 0.2464 \\
        & Random & +1.42 & 0.7396 \\
        & Representativeness & +2.43 & 0.4560 \\
        & Uncertainty & +6.91 & 0.0550 \\
        \midrule
        \multirow{5}{*}{\textbf{RandomForest}} 
        & Diversity & +2.66 & 0.3309 \\
        & Outlierness & +3.55 & 0.2441 \\
        & Random & +6.52 & 0.0505 \\
        & \textbf{Representativeness} & \textbf{+11.56} & \textbf{0.0004*} \\
        & \textbf{Uncertainty} & \textbf{+8.26} & \textbf{0.0331*} \\
        \midrule
        \multirow{5}{*}{\textbf{XGBoost}} 
        & Diversity & +0.95 & 0.6802 \\
        & Outlierness & +1.70 & 0.3492 \\
        & Random & +4.27 & 0.1283 \\
        & Representativeness & +0.89 & 0.7597 \\
        & Uncertainty & -0.35 & 0.8563 \\
        \bottomrule
    \end{tabular}
\end{table}

\newpage

\subsubsection{Starkey}
The \textit{Starkey} dataset serves as a control case for sufficient, high-quality data, with baselines exceeding 86\%. Table \ref{tab:final_analysis_starkey} illustrates the potential risks of augmentation in such environments.

For the Random Forest model, applying the Outlierness and Representativeness strategies actually resulted in statistically significant performance degradation (-0.95\% and -1.16\% respectively). This indicates that when the data manifold is already well-sampled, adding synthetic noise can corrupt the feature space rather than enhance it.

However, the MLP model again demonstrated a unique synergy with the Uncertainty strategy, achieving a statistically significant improvement of +3.18\% ($p = 0.0087$). This reinforces the finding from the AIS dataset that, while tree-based models may reject augmentation on high-quality data, neural networks can still benefit significantly from active-learning-style data injection to resolve specific ambiguities in the decision boundary.

% Auto-generated for starkey with 20 seeds.
\begin{table}[ht]
    \centering
    \caption{
        Statistical analysis of the performance improvement for each optimally-tuned 
        augmentation strategy compared to the baseline. This table presents the final, 
        rigorous comparison, in contrast to the exploratory overview in the previous table.
        An asterisk (*) denotes a statistically significant result (p < 0.05).
    }
    \label{tab:final_analysis_starkey}
    \begin{tabular}{{l l r r}}
        \toprule
        \textbf{Model} & \textbf{Strategy} & \textbf{Mean Improv. (\%)} & \textbf{p-value} \\
        \midrule
        \multirow{5}{*}{\textbf{LogisticRegression}} 
        & Diversity & +0.29 & 0.6303 \\
        & Outlierness & +0.75 & 0.2118 \\
        & Random & +0.52 & 0.2811 \\
        & Representativeness & -0.31 & 0.5700 \\
        & Uncertainty & +0.34 & 0.5900 \\
        \midrule
        \multirow{5}{*}{\textbf{MLP}} 
        & Diversity & +1.98 & 0.1031 \\
        & Outlierness & +1.15 & 0.3267 \\
        & Random & +2.07 & 0.1244 \\
        & Representativeness & +1.95 & 0.1005 \\
        & \textbf{Uncertainty} & \textbf{+3.18} & \textbf{0.0087*} \\
        \midrule
        \multirow{5}{*}{\textbf{RandomForest}} 
        & Diversity & -0.61 & 0.2526 \\
        & Outlierness & -0.95 & 0.0108 \\
        & Random & -0.36 & 0.3650 \\
        & Representativeness & -1.16 & 0.0414 \\
        & Uncertainty & -0.47 & 0.3742 \\
        \midrule
        \multirow{5}{*}{\textbf{XGBoost}} 
        & Diversity & +0.30 & 0.5614 \\
        & Outlierness & -0.12 & 0.8368 \\
        & Random & -0.20 & 0.6737 \\
        & Representativeness & -0.10 & 0.7993 \\
        & Uncertainty & +0.09 & 0.8744 \\
        \bottomrule
    \end{tabular}
\end{table}

\newpage

\subsection{Geometric Analysis via UMAP}
\label{sec:geometric_analysis}

The quantitative results in Section \ref{sec:overallresults} demonstrate that augmentation works in specific contexts. 
However, numbers alone do not explain why. 
To understand the underlying mechanisms driving these results, we visualize the impact of synthetic data on the feature manifold. The following analysis examines three distinct geometric phenomena observed during the experiments to illustrate how different strategies interact with the data topology.

\subsubsection{Visualization Methodology}

Directly visualizing the 72 dimensional feature space is impossible. Furthermore, standard reduction techniques such as principal component analysis (PCA) rely on linear projections that often obscure the complex boundaries characteristic of trajectory data.

Therefore, we employ Uniform Manifold Approximation and Projection (UMAP). UMAP constructs a high dimensional graph representation of the data using k Nearest Neighbors (kNN) and optimizes a low dimensional layout to preserve the local connectivity of that graph. This makes it uniquely successful in preserving the local density structure of the classes. It allows us to observe whether the augmented trajectories remain topologically connected to their parent clusters or drift into invalid regions.

To visualize the decision landscape within this 2D projection, we train a kNN classifier on the 2D embeddings. The background colors in the following figures represent the decision regions learned by this proxy classifier. 
This choice is methodological. Since UMAP is a neighbor-based algorithm, a kNN decision boundary provides the most faithful representation of how the local data density has been altered by augmentation.

% \textit{Note: The F1 Scores reported in the figure titles correspond to the Random Forest model trained on the full 72 dimensional dataset from Section \ref{sec:overallresults}. They do not correspond to the 2D proxy model. The visualizations serve to explain the geometric cause of these performance shifts.}

In the following visualizations\footnote{The F1 Scores reported in the figure titles correspond to the Random Forest model trained on the full 72 dimensional dataset from Section \ref{sec:overallresults}. They do not correspond to the 2D proxy model. The visualizations serve to explain the geometric cause of these performance shifts.}, the left panel represents the baseline model state and the right panel represents the state after retraining on augmented data:
\begin{itemize}
    \item \textbf{Circles ($\circ$)}: Original test data points.
    \item \textbf{Stars ($\star$)}: Trajectories selected for augmentation.
    \item \textbf{Pluses ($+$)}: New synthetic trajectories generated.
    \item \textbf{Dashed Lines}: Connect a parent trajectory to its augmented child. This visualizes the magnitude of the perturbation.
    \item \textbf{Background Color}: The decision region learned by the visualization proxy classifier (k Nearest Neighbors).
\end{itemize}

\newpage

% \subsubsection{Mechanism 1: Manifold Densification}
\subsubsection{Manifold Densification in the Foxes Dataset}

\begin{figure}[ht]
    \centering
    \includegraphics[width=\linewidth]{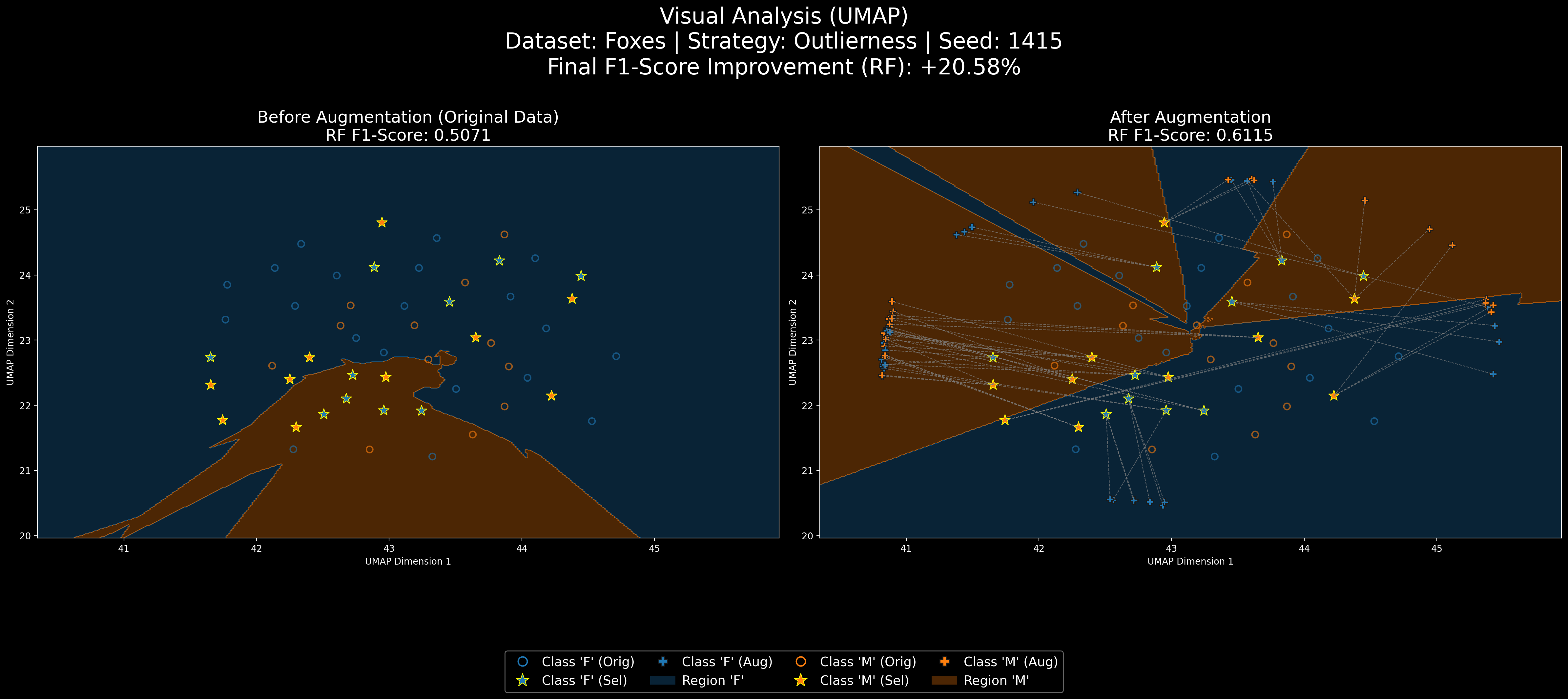}
    \caption{Visual analysis of Manifold Densification (Foxes, Seed 1415). The Outlierness strategy identifies sparse, fragmented regions of the minority class. By injecting synthetic density (pluses) into these gaps, the model is forced to carve out new decision regions (Right), significantly improving recall.}
    \label{fig:umap_foxes}
\end{figure}

The strongest validation of systematic selection is observed in the \textit{Foxes} dataset, where data is sparse and noisy. Figure \ref{fig:umap_foxes} illustrates a specific seed (1415) in which the Outlierness strategy improved the Random Forest F1-score by +20.58\%.

\noindent Visual inspection of the baseline state (Left) reveals a highly fragmented manifold. The decision boundary (dark blue vs. brown) is simplistic, failing to capture the complex, scattered nature of the classes. Specifically, the model misses several sub-clusters of the minority class, merging them into the background.

The Outlierness strategy (Right) successfully identifies these boundary cases, namely trajectories that lie at the edges of the known distribution. The augmentation process generates clouds of synthetic points (pluses) around these selected outliers. This process, which we term Manifold Densification, artificially increases the local density of the minority class in sparse regions.

The impact on the classifier is drastic: the decision boundary shifts from a simple curve to a complex, multi-modal shape. Note the large new brown region that appears in the top-right quadrant of the augmented plot. The model has \enquote{learned} a new valid region of the feature space that was previously ignored, explaining the massive +20\% performance jump.

\newpage

% \subsubsection{Mechanism 2: Decision Margin Corruption}

\subsubsection{Decision Margin Corruption in the Starkey Dataset}

\begin{figure}[ht]
    \centering
    \includegraphics[width=\linewidth]{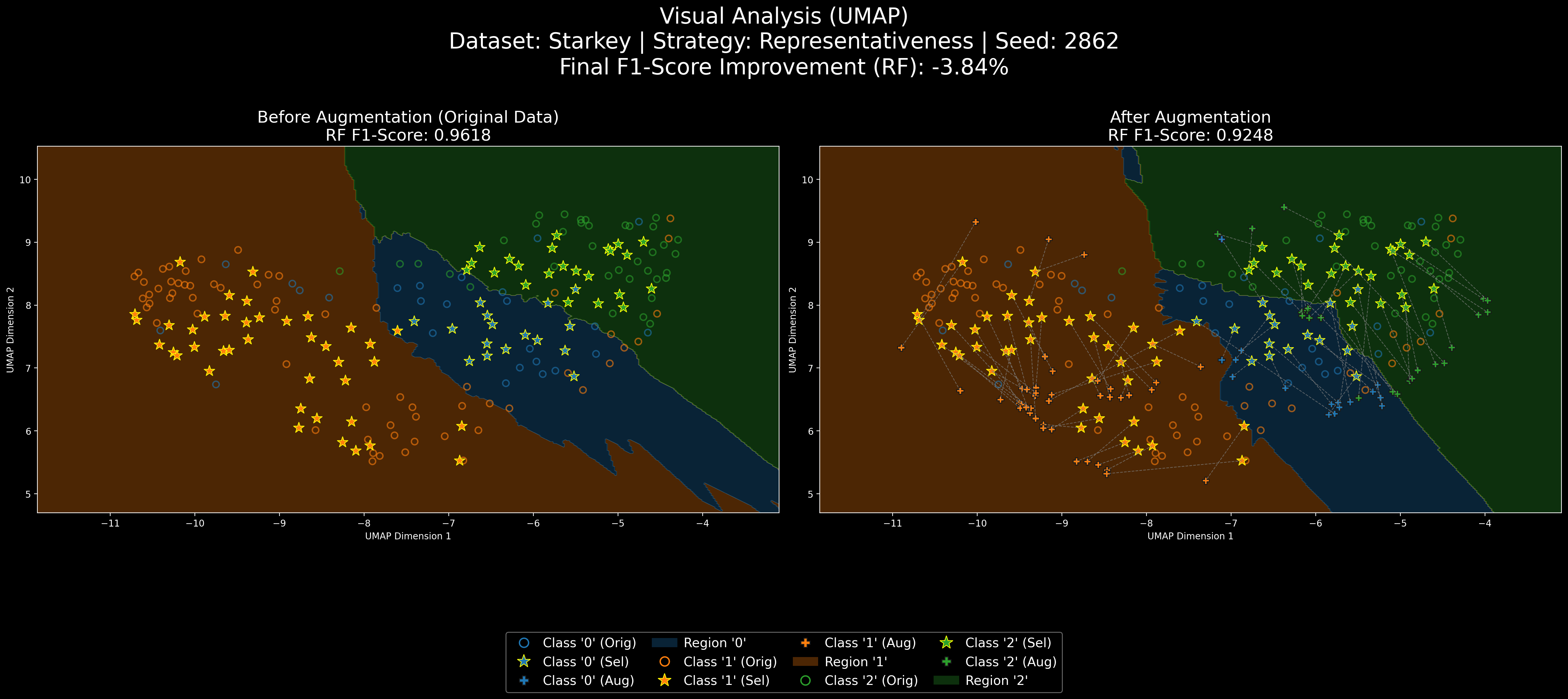}
    \caption{Visual analysis of Decision Margin Corruption (Starkey, Seed 2862). The Representativeness strategy selects central, typical examples (stars). However, in a dense dataset, adding noise to these points blurs the separation between classes (Right), causing the decision boundary to become jagged and less precise.}
    \label{fig:umap_starkey}
\end{figure}

Although augmentation is beneficial for sparse data, Figure \ref{fig:umap_starkey} demonstrates why it can fail on high-quality data. In the \textit{Starkey} dataset (Seed 2862), the Representativeness strategy caused a performance degradation of -3.84\%.

\noindent In the baseline state (Left), the classes are well-separated with high density (F1-Score $>$ 96\%). The decision boundaries are sharp and follow the natural gaps between clusters. The Representativeness strategy selects the most \enquote{typical} points near the centroids of these clusters (indicated by stars deep inside the class blobs).

However, the augmentation process injects noise into these central points, spreading them outward toward the class boundaries (indicated by the dashed lines pushing outwards). Instead of reinforcing the class definition, this corrupts the decision margin. The synthetic points fill the clean gaps that previously separated the classes. 

The Random Forest, attempting to classify this noisy overlap, produces a highly irregular decision boundary with excessive complexity (Right). This over-fitting to synthetic noise reduces the model's ability to generalize to the clean test set, resulting in a net loss of accuracy. This confirms that when a manifold is already well-defined, heuristic augmentation often acts as a contaminant rather than a regularizer.

It is critical to highlight that this performance degradation is not a failure of the experimental methodology, but rather a primary finding of this research. By systematically observing how noise is added to the natural differences between classes, we demonstrate that geometric augmentation is fundamentally inappropriate for clean, well-separated manifolds, as it forces the model to overfit to an artificial overlap.

\newpage

\subsubsection{Feature Space Divergence in Car Dataset}
\begin{figure}[ht]
    \centering
    \includegraphics[width=\linewidth]{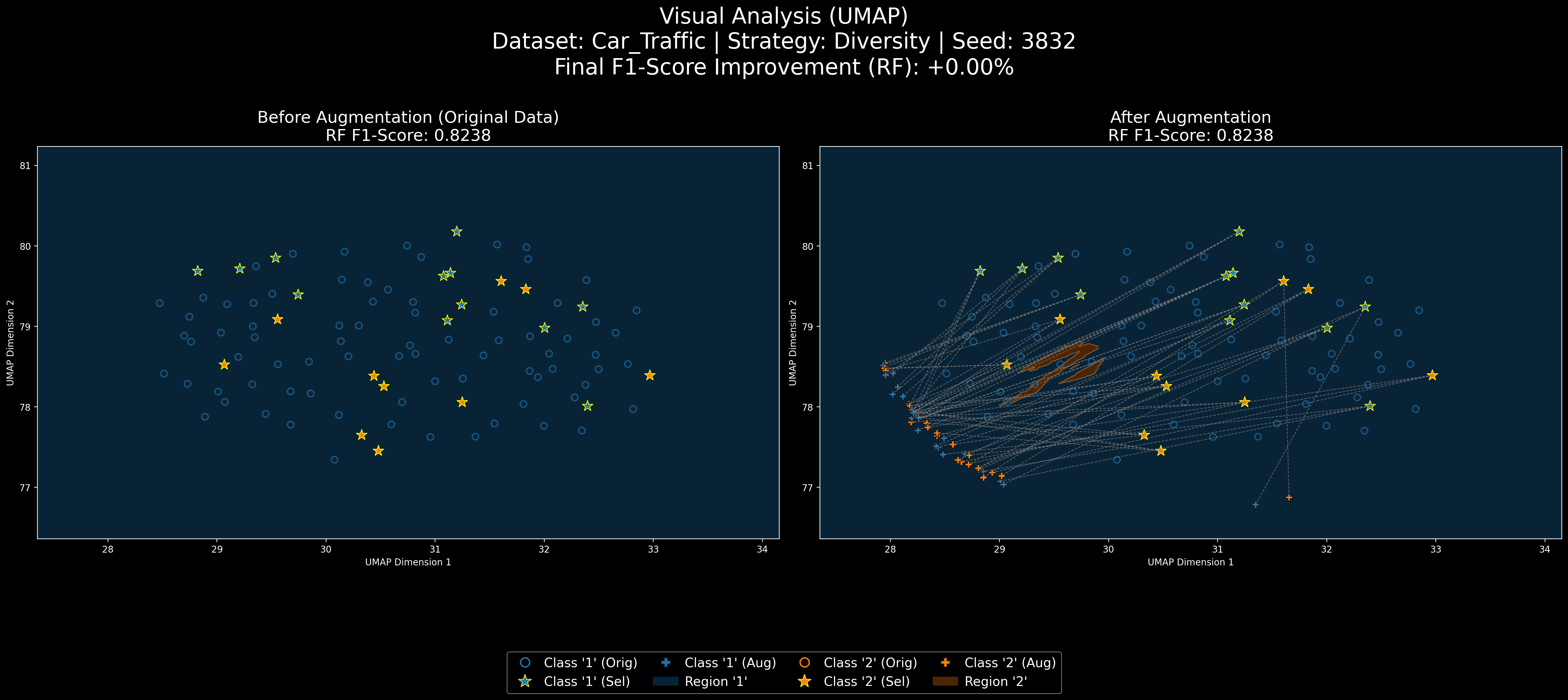}
    \caption{Visual analysis of Feature Space Divergence (Car Traffic, Seed 3832). Due to high object velocity, the geometric perturbation projects augmented points far away from the original manifold (Right). The model learns to classify these \enquote{out-of-distribution points} (pluses on the far left), but the decision boundary around the real data remains unchanged.}
    \label{fig:umap_traffic}
\end{figure}

The final case study explains the \enquote{null result} often observed in the Car Traffic dataset. Figure \ref{fig:umap_traffic} shows Seed 3832 under the Diversity strategy, where the net performance improvement was exactly +0.00\%.

\noindent This visualization captures a phenomenon driven by the physics of the domain. Because cars travel at high velocities, the physical distance between GPS updates is large. Consequently, a percentage-based perturbation (e.g., 50\% of the segment length) results in a massive geographical displacement.

Because the algorithm does not take into account the existence of absolute physical constraints (such as road networks), the creation of these artificial paths results in trajectories with a high chance of passing through buildings or off-road areas. As observed by the 0.00\% improvement, this lack of physical plausibility renders the augmented data semantically useless to the classifier in urban contexts.

In the UMAP projection (Right), this is visible as Feature Space Divergence. The augmented points (pluses) are projected far to the left, completely disconnected from the original data cluster (circles). The dashed lines connecting parents to children are extremely long, indicating that the synthetic data no longer resemble the original distribution therefore it is \enquote{out-of-distribution}.

The classifier successfully learns to classify these new \enquote{out-of-distribution} clusters, creating complex decision regions on the far left of the plot. However, because these new regions do not intersect with the original data manifold, the decision boundary \textit{within} the dense test cluster remains virtually identical to the baseline. As a result, the classification of the test set is unchanged, leading to the observed 0.00\% improvement. This highlights a critical physical limit of geometric augmentation for high-velocity domains.

\newpage

\subsection{Impact of Model Architecture}
\label{sec:model_impact}

The impact of augmentation varied significantly depending on the machine learning algorithm used. The results highlight how different model architectures react to synthetic noise.

\begin{itemize}
    \item \textit{Linear Models (Logistic Regression):} These models showed the most binary response. In the sparse \textit{Foxes} dataset, where the linear decision boundary was initially poor, augmentation provided the largest relative gains by reshaping the class distribution. However, in dense datasets like \textit{Car Traffic}, these models reached saturation. Once the best possible linear boundary was found using the original data, adding synthetic points provided no benefit. This confirms that linear models are limited by their own simplicity rather than just data volume in high density scenarios.

    \item \textit{Tree Ensembles (Random Forest, XGBoost):} These models generally had the highest baseline scores, but also showed the highest sensitivity to bad augmentation. In the \textit{Starkey} dataset, Random Forest performance was significantly degraded when using the Representativeness strategy (-1.16\%). This indicates that tree based models, which create complex decision boundaries by splitting feature space, are prone to overfitting the synthetic noise when it fills the gaps between classes.

    \item \textit{Neural Networks (MLP):} The MLP showed a specific robustness in datasets where tree based models plateaued. In both the \textit{AIS Subset} and \textit{Starkey} datasets, the Random Forest model failed to improve or degrade. In contrast, the MLP achieved statistically significant gains using the Uncertainty strategy (+1.61\% and +3.18\% respectively). Although this trend was not universal across all datasets (Random Forest outperformed MLP in \textit{Foxes}), it suggests that the neural network was better able to utilize the ambiguous examples selected by the Uncertainty strategy to refine its probability estimates in dense data regimes.
\end{itemize}

\subsection{Summary of Findings}
\label{sec:summary_findings}

The evaluation of systematic trajectory selection strategies reveals four main findings that address the central research objectives of this thesis:

\begin{itemize}
    \item \textit{Systematic Selection Reduces Risk:} While systematic strategies did not always guarantee a performance boost, they proved generally more robust than the Random baseline. The Random strategy frequently resulted in performance degradation, particularly in the \textit{Starkey} and \textit{AIS} datasets. Systematic strategies like Outlierness and Uncertainty either provided gains or maintained parity, effectively acting as a more robust alternative to naive sampling.

    \item \textit{Augmentation is Conditional on Data Scarcity:} The results define a clear boundary for when augmentation is useful. In the data scarce \textit{Foxes} dataset, every strategy worked, yielding gains. In the data rich \textit{Starkey} dataset, most strategies failed or degraded performance. This proves that geometric augmentation acts primarily as a regularizer: it is transformative when data is lacking, but can become a source of corrupting noise when the dataset is already sufficient.

    \item \textit{Uncertainty Selection Aids in Dense Environments:} In scenarios where the dataset was large and dense (AIS and Starkey), standard geometric heuristics often failed. The Uncertainty strategy was the only approach to yield statistically significant improvements in these difficult conditions, specifically for the MLP model. This indicates that targeting the specific trajectories that confuse the model is more effective than targeting geometric outliers when the data manifold is already well populated.

    \item \textit{Physical Limits of Percentage-Based Noise:} The results from the \textit{Car Traffic} dataset reveal a technical limitation. The standard approach of defining the perturbation radius as a percentage of segment length breaks down for high velocity objects. This resulted in synthetic data being projected far outside the valid feature space (Feature Space Divergence), rendering the augmentation mathematically valid but practically useless for the classifier. This \enquote{null} finding confirms that standard geometric augmentation procedures are essentially out of phase with high-velocity realms and require map-matching constraints to be viable.
\end{itemize}

\newpage

\section{Discussion}
\label{sec:discussion}

This thesis aims to answer a fundamental question: Can trajectories be systematically selected for augmentation to improve classification performance? 
The preceding results provide a nuanced and conditional answer. The findings indicate that while systematic selection strategies generally offer a more robust alternative to random sampling, they do not universally improve classification performance over the original baseline in all contexts. Instead, the utility of these strategies is primarily determined by the quality and density of the underlying data. While systematic approaches deliver improvements in sparse and noisy regimes, they often reach a saturation point in high quality datasets where the decision boundaries are already well defined. This section interprets these findings, discusses their implications in the context of the existing literature, and contextualizes the contributions of this work.

\subsection{The Primacy of Selection: An Empirical Answer to a Known Gap}
A primary finding of this study is that systematic selection strategies generally offer distinct advantages over a naive random approach. Although random selection is often volatile and prone to inducing performance degradation, systematic strategies provide a more stable and effective method for identifying augmentation candidates. The results demonstrate that the strategic selection of data is a critical determinant of augmentation success, especially in data scarce regimes.The foundational work by Haranwala et al.~\cite{haranwala2023data} first demonstrated the viability of perturbation based augmentation but explicitly acknowledged that their reliance on purely random selection was a key limitation, noting that \enquote{innovating new techniques for ranking trajectories presents an exciting possibility.} The findings in this thesis provide a direct and empirical answer to the gap they identified by demonstrating that selection logic is a critical factor for augmentation success.

Although the later AugmenTRAJ framework~\cite{haranwala2023augmenttraj} began to address this by offering several predefined heuristic strategies, this work goes a step further. By providing a data driven comparison of strategies grounded in core statistical properties (outlierness, diversity, representativeness, uncertainty), it confirms that how trajectories are selected determines the stability of the model. Simply increasing the volume of training data is insufficient; the strategic choice of which examples to generate is paramount to avoid the volatility and performance degradation frequently observed in the random baseline. The goal of these systematic strategies is not to claim universal performance dominance, but rather to mitigate risk. When the standard method of augmenting data (random sampling) is highly volatile, applying a systematic approach ensures that we do not unknowingly degrade the model's original base performance.

\subsection{Interpreting the Strategy Mechanisms in Context}
The varied performance of the systematic strategies offers deep insight into their underlying mechanisms and allows for a comparison with similar concepts in the literature. The Outlierness based selection proved to be one of the most reliable performer in sparse regimes, acting as a \enquote{boundary refiner.} This approach validates the practical implementation of the idea that the most informative samples for a classifier are often those near the decision boundary. By augmenting the most unusual examples, this strategy effectively forces the model to expand its decision region to encompass valid but rare trajectory patterns.

Diversity based selection acts as a \enquote{feature space explorer.} Its use of clustering to identify and sample from various groups of trajectories shares a philosophical similarity with the work of Mirkhani et al.~\cite{mirkhani2024augmenting}, who used clustering to identify and augment safety critical driving scenarios. However, our results add a caveat to this approach: exploring the feature space is beneficial only when the augmentation mechanism respects the domain physics. In the \textit{Car Traffic} dataset, diversity maximization led to feature space divergence, creating synthetic data that are too distinct from the original manifold to be useful.

Representativeness based selection offered diminishing returns or degradation in high quality datasets. By reinforcing the \enquote{average} patterns, this strategy appears to compound the density at the cluster centers. In dense datasets like \textit{Starkey}, this outward expansion from the center resulted in the corruption of decision margins. This suggests that for standard classification, the most valuable information for a model often lies not at the class centroid, but at its boundaries.

Uncertainty based selection demonstrated a unique synergy with neural network architectures. While tree based models often rejected augmentation in dense regimes, the MLP model extracted value from the ambiguous examples selected by this strategy. This effectively transfers the benefits of Active Learning to the augmentation domain, suggesting that for non convex optimization problems, targeting the specific instances that confuse the model is more robust than relying on geometric heuristics alone.

A notable observation across all datasets is the absence of a single \enquote{winning}strategy that is universally effective. The results demonstrate a deep sensitivity between the selection strategy and the underlying model architecture. For instance, the Uncertainty strategy frequently assisted the Multi-Layer Perceptron (MLP) by resolving ambiguous boundary cases, but it had little to no positive effect on tree-based ensembles like Random Forest on the same data. This inconsistency implies that selection strategies must be carefully mapped to the learning dynamics of the chosen classifier, a complex relationship that warrants deeper investigation in future research.

\subsection{The Conditional Value of Augmentation}
Perhaps the most crucial contribution of this thesis is the empirical demonstration that data augmentation is not a universal panacea, but a powerful tool whose value is conditional. The contrasting results from the \textit{Starkey} and \textit{Foxes} datasets provide clear evidence for this principle.

The performance degradation on the high quality \textit{Starkey} dataset reveals a critical limitation of simple perturbation techniques: when applied to sufficient, well separated data, augmentation acts as a harmful \enquote{corrupting noise signal.} This finding serves as an important counterpoint to the generally positive results reported by Haranwala et al.~\cite{haranwala2023data} and highlights a critical prerequisite for applying these methods. Augmentation functions best as a repair mechanism for data deficiency rather than as a generic performance booster.

\subsection{Physical Constraints and Domain Dynamics}
The null results observed in the \textit{Car Traffic} dataset highlight a critical dependency between the augmentation algorithm and the physical dynamics of the moving object. The standard approach of defining the perturbation radius as a percentage of the segment length implicitly assumes spatial continuity. However, high velocity objects create sparse, long distance segments. Applying a percentage based perturbation to these segments results in \enquote{Kinematic Distortion}, projecting points far away from the road network.

This aligns with challenges seen in other domains, such as the work by Antotsiou et al.~\cite{antotsiou2021adversarial} who sought to preserve the validity of augmented actions in imitation learning. Future frameworks must incorporate absolute physical constraints rather than rely solely on relative geometric ratios.

\subsection{Implications of Per Dataset Optimization}
A final critical contribution of this work is the validation of the hyperparameter optimization strategy per data set. This approach goes beyond the static heuristics of AugmenTRAJ~\cite{haranwala2023augmenttraj} and introduces a more advanced adaptive methodology. The fact that the Optuna framework identified vastly different optimized configurations for each dataset is a finding in its own right. The sparse \textit{Foxes} dataset required aggressive augmentation (40\% selection, 5 variations), while the dense \textit{Car Traffic} dataset tolerated only minimal intervention.

This aligns with the principles of sophisticated selection mechanisms in other domains, such as SelectAugment for image data~\cite{lin2023selectaugment}, which advocate learning \textit{which} and \textit{how} to augment rather than relying on fixed rules. By integrating Optuna, this thesis provides a methodological blueprint for applying this data driven philosophy to the trajectory domain.

\section{Conclusion and Future Work}
\label{sec:future_work}

This thesis established a systematic and scalable framework for selecting specific trajectories for data augmentation. By comparing strategies based on outlierness, diversity, representativeness, and prediction uncertainty, the research moved beyond simple random sampling to provide a more controlled approach to mobility analysis. The results show that systematic selection is generally safer and more stable than random selection, although its success depends heavily on the dataset. The findings confirm that augmentation is most effective as a repair mechanism for sparse or fragmented data. In dense and high quality datasets, even systematic augmentation can act as a corrupting noise signal that reduces model accuracy by blurring the decision margins between classes.

The research also identified a clear physical limit to geometric augmentation in high velocity domains. The results from the car traffic data demonstrated that percentage based perturbations can lead to feature space divergence, where synthetic trajectories become mathematically valid but practically useless because they no longer resemble real movement patterns. These insights show that the strategic selection of data is not just a technical detail but a primary factor in whether augmentation succeeds. This work provides a practical guide for using an optimized tuning process to identify exactly when and how to apply synthetic data to improve machine learning models. \\

For future research, the following areas should be explored:
\begin{enumerate}
\item \textit{Tuning the perturbation radius:} This study used a fixed radius of 50 percent for the noise injection. Future work should include this radius as a parameter in the optimization loop. Letting a tool like Optuna choose a smaller radius for high quality datasets might prevent the performance degradation observed in this research and allow for more subtle improvements.
\item \textit{Adding road network constraints:} To fix the divergence seen in vehicle trajectories, future algorithms should incorporate spatial constraints. This would ensure that synthetic points are shifted only to locations that exist on valid road networks or paths, preserving the semantic meaning of the movement in urban environments.
\item \textit{Testing other selection metrics:} While this study tested five core strategies, other methods for ranking trajectories could be investigated. Future research could explore selection based on the geometric complexity of a path or the specific contribution of a trajectory to the overall class boundary, which may provide even better results for linear models.
\end{enumerate}

%----------------------------------------------------------------------------------------
%	References. IEEE style is used.
%
%----------------------------------------------------------------------------------------
\newpage

\hypersetup{urlcolor=black}
\bibliographystyle{IEEEtran}
\bibliography{referenser.bib}

\newpage
%----------------------------------------------------------------------------------------
%	Appendix
%-----------------------------------------------------------------------------------------
\pagenumbering{Alph}
\setcounter{page}{1} % Reset page numbering for Appendix
\appendix

\section*{Appendix 1: Source Code and Reproducibility}
To ensure full transparency and reproducibility of the results discussed in this thesis, the complete Python framework has been open-sourced. The repository includes the data preprocessing scripts, the systematic selection algorithms, the multi-core feature extraction optimized with Numba, and the Optuna hyperparameter optimization database. \\

\noindent \textbf{GitHub Repository:} {\small \url{https://github.com/adamnordling/master-thesis-trajectory-data-augmentation}}
%In the appendix, you can add details that do not fit into the main report. 

%Examples could be source code, long tables with raw data, questionnaires, questionnaires raw data, etc.
\end{document}